\documentclass{article} 
\usepackage{iclr2026_conference,times}

\newcommand{\addcite}[1]{\textbf{\color{blue}}}

\definecolor{commentblue}{RGB}{30,90,160}

\usepackage{amsmath,amsfonts,bm}

\def\eqref#1{equation~\ref{#1}}

\def\1{\bm{1}}

\def\vs{{\bm{s}}}

\DeclareMathAlphabet{\mathsfit}{\encodingdefault}{\sfdefault}{m}{sl}
\SetMathAlphabet{\mathsfit}{bold}{\encodingdefault}{\sfdefault}{bx}{n}

\usepackage{hyperref}
\usepackage{url}
\usepackage{xspace}
\usepackage{xcolor}     
\usepackage{colortbl}  
\usepackage{subcaption}
\usepackage{booktabs} 
\usepackage{graphicx}
\usepackage{multicol}
\usepackage{multirow}
\usepackage{wrapfig}
\usepackage{pifont}
\usepackage{diagbox}
\usepackage{tikz}
\usepackage[most]{tcolorbox}
\usepackage[dvipsnames]{xcolor}
\usepackage[table]{xcolor}
\usepackage[capitalize]{cleveref}
\crefname{section}{Sec.}{Secs.}
\Crefname{section}{Section}{Sections}
\Crefname{table}{Table}{Tables}
\crefname{table}{Tab.}{Tabs.}
\usepackage{tcolorbox}
\tcbuselibrary{skins}
\definecolor{headergray}{RGB}{60, 60, 60} 
\definecolor{customblue}{HTML}{4A90E2}
\usetikzlibrary{shapes}

\newcommand\mypara[1]{\vspace{0mm}\noindent\textbf{#1}}

\newcommand{\tablestyle}[2]{\setlength{\tabcolsep}{#1}\renewcommand{\arraystretch}{#2}\centering\footnotesize}
\newlength\savewidth

\usepackage{tabularx} 

\definecolor{mycyan}{cmyk}{.1,0,0,0}

\newcommand{\fullcheckmark}{\textcolor{ForestGreen}{\ding{51}}}

\makeatletter
\DeclareRobustCommand\onedot{\futurelet\@let@token\@onedot}
\def\@onedot{\ifx\@let@token.\else.\null\fi\xspace}

\def\eg{\emph{e.g}\onedot} 
\def\ie{\emph{i.e}\onedot} 
 
 \def\vs{\emph{vs}\onedot}

\makeatother

\title{VLNVerse: A Benchmark for Vision-Language Navigation with Versatile, Embodied, Realistic Simulation and Evaluation}

\author{\textbf{Sihao Lin}$^{1,2}$\thanks{Equal Contribution, $\dagger$ Corresponding Author.} \quad 
\textbf{Zerui Li}$^{1,2*}$ \quad 
\textbf{Xunyi Zhao}$^{1,2*}$ \quad 
\textbf{Gengze Zhou}$^1$\\
\textbf{Liuyi Wang}$^3$ \quad 
\textbf{Rong Wei}$^4$ \quad 
\textbf{Rui Tang}$^4$ \quad 
\textbf{Juncheng Li}$^5$ \quad
\textbf{Hanqing Wang}$^6$ \\ 
\textbf{Jiangmiao Pang}$^6$ \quad 
\textbf{Anton van den Hengel}$^{1,2}$ \quad 
\textbf{Jiajun Liu}$^{2,7}$ \quad 
\textbf{Qi Wu}$^{1,2\dagger}$\\
$^1$Adelaide University; $^2$Responsible AI Research Centre, Australian Institute for Machine Learning; \\ $^3$Tongji University; $^4$ ManyCore; $^5$ Zhejiang University; $^6$ Shanghai AI Lab; $^7$CSIRO Data61\\
\texttt{\{sihao.lin,zerui.li,xunyi.zhao,gengze.zhou,qi.wu01\}@adelaide.edu.au} \\
}

\iclrfinalcopy 
\begin{document}

\maketitle

\begin{center}
    \vspace{-20pt} 
    
    \href{https://sihaoevery.github.io/vlnverse/}{\color{magenta} https://sihaoevery.github.io/vlnverse/}
    
    \vspace{5pt} 
\end{center}

\begin{abstract}
Despite remarkable progress in Vision-Language Navigation (VLN), existing benchmarks remain confined to fixed, small-scale datasets with naive physical simulation. These shortcomings limit the insight that the benchmarks provide into sim-to-real generalization, and create a significant research gap. Furthermore, task fragmentation prevents unified/shared progress in the area, while limited data scales fail to meet the demands of modern LLM-based pretraining. To overcome these limitations, we introduce VLNVerse: a new large-scale, extensible benchmark designed for \textbf{V}ersatile, \textbf{E}mbodied, \textbf{R}ealistic \textbf{S}imulation, and \textbf{E}valuation. VLNVerse redefines VLN as a scalable, full-stack embodied AI problem. Its \textit{Versatile} nature unifies previously fragmented tasks into a single framework and provides an extensible toolkit for researchers. Its \textit{Embodied} design moves beyond intangible and teleporting ``ghost" agents that support full-kinematics in a \textit{Realistic Simulation} powered by a robust physics engine. We leverage the scale and diversity of VLNVerse to conduct a comprehensive \textit{Evaluation} of existing methods, from classic models to MLLM-based agents. We also propose a novel unified multi-task model capable of addressing all tasks within the benchmark. VLNVerse aims to narrow the gap between simulated navigation and real-world generalization, providing the community with a vital tool to boost research towards scalable, general-purpose embodied locomotion agents.

\end{abstract}
 
\section{Introduction}
\label{sec:intro}

Vision-Language Navigation (VLN)~\citep{anderson2018r2r} is a fundamental task in Embodied AI where an agent must comprehend natural language instructions to navigate through a 3D space and arrive at a target. VLN is essential for developing versatile Embodied AI, as it seamlessly combines multi-modal data comprehension with action prediction in dynamic settings, making it an excellent platform to evaluate the entire perception-to-action pipeline~\citep{nguyen2021sensorimotor, duan2022survey, habitat2, hu2023_uniad,Chu2023MobileVLMA}. Conventionally, VLN models are often trained and evaluated in simulated environments ~\citep{misra2018mapping, anderson2020rxr, jain2019stay, nguyen2019vision, irshad2021robovln, liu2023aerialvln, qi2020reverie, thomason2020cvdn, song2025lhvln, nguyen2019help, krantz2020beyond, shridhar2020alfred, padmakumar2022teach, gao2022dialfred, zhao2023mind}. 
However, the advancement of VLN is fundamentally constrained by the limitations of existing simulation-based platforms. While current simulators have laid the groundwork, they exhibit significant shortcomings that impede progress toward generalizable and deployable agents. For instance, widely-used environments like Matterport3D (MP3D)~\citep{chang2017matterport3d} are built on static and discrete graphs, effectively degenerating agents to ``ghost-like'' entities devoid of physical properties. Subsequent platforms like Habitat~\citep{savva2019habitat} and Gibson~\citep{xia2018gibson} have introduced physics, yet they often rely on simplified engines with limited support for complex robot kinematics, continuous control, and flexible user customization. This lack of a high-fidelity simulation represents a critical bottleneck for the field.

This fragmentation at the simulator level directly propagates to task design and algorithmic development. The VLN landscape is populated with disparate benchmarks~\citep{anderson2018r2r, anderson2020rxr, jain2019stay, qi2020reverie, thomason2020cvdn, song2025lhvln, nguyen2019help}, each optimized for the specific capabilities of its underlying simulator. Naively unifying these disparate tasks on existing platforms is often impractical, as the simulators themselves are not designed for the general physics support required for true versatility. Consequently, existing algorithms are often over-specialized. That is, a model developed for one dataset (\eg, R2R~\citep{anderson2018r2r}) frequently requires non-trivial modifications to be applied to another, obstructing direct comparison and the knowledge transfer. This paradigm of task-specific solutions fundamentally contradicts the goal of general-purpose embodied agents.

\begin{figure*}[t]
    \centering
    \includegraphics[width=\textwidth]{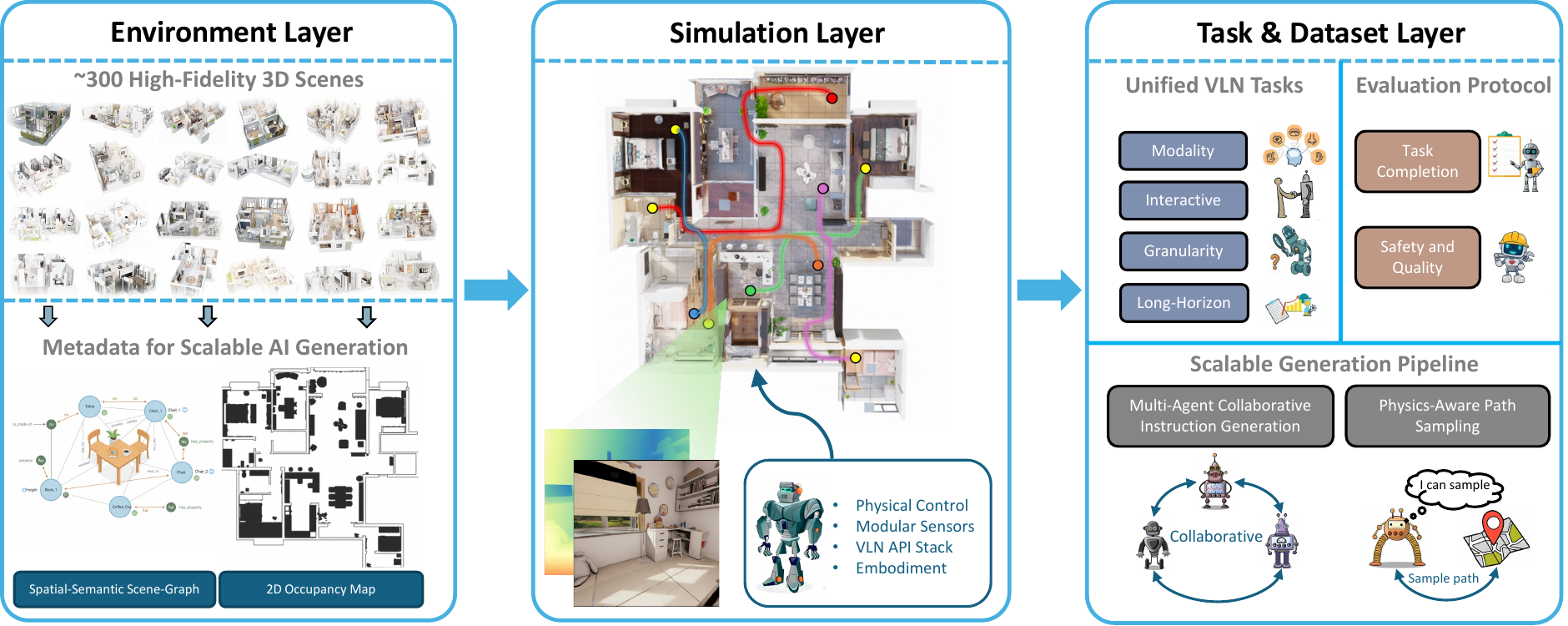}
    \caption{VLNVerse is built on a decoupled three-layer architecture that separates the responsibilities of the Agent (Simulator Layer)~\cref{sec:sim}, World (Environments Layer)~\cref{sec:assets}, and Benchmark (Task \& Dataset Layer)~\cref{sec:data_pipeline}.}
    \label{fig:intro}
\end{figure*}

The root of this fragmentation lies in the absence of truly interactive and physics-aware 3D environments. Current benchmarks are predominantly built upon static 3D scans or environments with only hard-coded and simplistic physical interaction~\citep{savva2019habitat,xia2018gibson}. They fail to support dynamic interactions between an embodied agent and its surroundings, where an embodied agent must obey the laws of physics and account for its own physical form (kinematics and dynamics)~\citep{ma2022dorothie, zhouhazard, puighabitat,li2024havln}. This deficiency creates a significant sim-to-real gap, as models trained in these abstract environments fail when confronted with the complexities of real-world robotics. Consequently, the field lacks a scalable and unified benchmark designed from the ground up to propel research towards physics-aware navigation and bridge the gap between simulation and deployment~\citep{qiao2024open, shi2025smartway, li2025ground, wang2024sim}.

These limitations are further compounded by issues of data scale and outdated training paradigms. Existing VLN datasets are often small-scale, inconsistent with the data-hungry trend of modern Multimodal Large Language Models (MLLMs)~\citep{wang2023scaling}. Furthermore, their static nature binds benchmarks to discrete environments and fixed modalities (\eg, RGBD), making it difficult to extend them to new sensory information or adapt to continuous action spaces. As a result, many advances are confined to offline training regimes, which further hinders physical interaction and obstructs sim-to-real deployment~\citep{hong2020recurrent, chen2021hamt, chen2022duet}.

To address this fundamental limitation, we propose VLNVerse, a new framework for VLN , specifically designed for \textbf{V}ersatile, \textbf{E}mbodied, \textbf{R}ealistic \textbf{S}imulation and \textbf{E}valuation. VLNVerse is a 
scalable \textbf{full-stack development} framework for VLN, covering embodiment, task fragment, data pipeline, and sim-to-real simulation. Compared to previous benchmark, we redefine VLN from the following aspects:

\mypara{High-Fidelity Physics-Aware Simulation.} We move beyond ``ghost-like" agents by building VLNVerse on NVIDIA Isaac Sim~\citep{isaacsim}, leveraging its robust physics engine and photorealistic rendering. Unlike previous simulators that treat navigation as discrete teleportation~\citep{anderson2018r2r}, we provide a comprehensive embodied robotics framework with unified APIs specifically tailored for VLN research. Specifically, we seamlessly integrate Isaac Sim's physics capabilities into VLN-specific workflows, eliminating the need for researchers to handle low-level robotics details. This foundation enables true embodiment, supporting robot kinematics, continuous control, collision dynamics, and physical object interactions. This design directly targets the sim-to-real gap by forcing agents to obey physical constraints.

\mypara{Unified Multi-Task Framework.} The versatility of our simulator allows us to unify the fragmented landscape of VLN  tasks. VLNVerse introduces a unified task taxonomy~\citep{zhou2024same} that covers a comprehensive spectrum of navigation challenges: (1) \textit{fine-grained navigation}, (2) \textit{coarse-grained (goal-oriented) navigation}, (3) \textit{visual-target search}, (4) \textit{long-horizon, multi-stage tasks}, and (5) \textit{interactive dialogue-based navigation}. This comprehensive taxonomy spans multiple dimensions: granularity, modality, temporal scope, and interaction type. Critically, VLNVerse unifies previously disparate benchmarks under a single framework, enabling the development and evaluation of general-purpose agents. The modular design also ensures effortless extensibility, where researchers can define and integrate new tasks by simply composing existing primitives.

\mypara{Scalable and Flexible Data Pipeline.}
To support modern data-hungry models and diverse training paradigms, we provide 263 large-scale, interactive 3D home environment assets. More importantly, we release a scalable data generation toolkit that shifts VLN from static, fixed-size datasets to a dynamic, on-demand generation paradigm. Such a pipeline enable generate training instances at arbitrary scales, customize environmental parameters, and define novel task configurations on demand. This pipeline flexibly supports both offline pre-training and online, interactive fine-tuning.

\mypara{Novel Unified Multi-Task Model.} To promote a shift away from over-specialized models, we propose a unified multi-task navigation model capable of handling all VLNVerse tasks simultaneously. Built on a State-Adaptive Mixture-of-Experts architecture~\citep{zhou2024same}, our model learns task-specific representations while enabling cross-task knowledge transfer. This approach serves as a baseline and represents a step towards general-purpose agent development.

\mypara{Systematic Evaluation and Baselines.}
As a new benchmark, we conduct a systematic evaluation of existing methods spanning the evolution of VLN, from classic Seq2Seq~\citep{sutskever2014sequence} models to cutting-edge MLLM-based agents (\eg, NavGPT~\citep{zhou2024navgpt}, MapGPT~\citep{chen2024mapgpt}). We re-implement and test these models in our physics-aware setting, providing crucial insights into their robustness and generalization under embodiment constraints. All implementations, datasets, and evaluation protocols will be made publicly available to ensure reproducibility and facilitate future research.

\section{Related Datasets}

\begin{table*}[t!]
    \centering
    \caption{Comprehensive comparison of VLN Datasets. 
    \textbf{Simulator}: `Specialized' implies physics restricted to specific cases; `Full' implies total physical simulation. \textbf{New Scenes}: Count of unique environments introduced (0 denotes directly re-use of existing scenes). 
    \textbf{Taxonomy}: F (Fine-grained), C (Coarse-grained), I (Interactive), LH (Long-horizon). 
    \textbf{Action Space}: `Discrete' (Discrete steps), `Graph' (Node-hopping), `Continuous' (Velocity based control), `Hybrid' (Multi-modal control).
    }
    \resizebox{\textwidth}{!}{
    \begin{tabular}{lccccccccc} 

\toprule
\multirow{2}{*}{\textbf{Dataset}} & \multicolumn{4}{c}{\textbf{Simulator}} & \multicolumn{4}{c}{\textbf{Taxonomy}} & \multirow{2}{*}{\textbf{Action Space}} \\ 
\cmidrule(lr){2-5} \cmidrule(lr){6-9}
& \textbf{Name} & \textbf{Physical} & \textbf{Render Quality} & \textbf{New Scenes \#} & \textbf{F} & \textbf{C} & \textbf{I} & \textbf{LH} & \\
\midrule 

LANI/CHAI~\citeyearpar{misra2018mapping} & CHALET & Specialized & Low & 1 & \fullcheckmark& -& -& -& Discrete \\

R2R~\citeyearpar{anderson2018r2r} & Materport3D & - & Low & 90 & \fullcheckmark& -& -& -& Graph \\

R4R~\citeyearpar{jain2019stay} & Materport3D & - & Low & 0 & \fullcheckmark& -& -& \fullcheckmark& Graph \\

RxR~\citeyearpar{anderson2020rxr} & Materport3D & - & Low & 0 & \fullcheckmark& -& -& -& Graph \\

REVERIE~\citeyearpar{qi2020reverie} & Materport3D & - & Low & 0 & -& \fullcheckmark& -& -& Graph \\

SOON~\citeyearpar{zhu2021soon} & Materport3D & - & Low & 0 & -& \fullcheckmark& -& -& Graph \\

VNLA~\citeyearpar{nguyen2019vision} & Materport3D & - & Low & 0 & -& -& \fullcheckmark& -& Graph \\

HANNA~\citeyearpar{nguyen2019help} & Materport3D & - & Low & 0 & -& -& \fullcheckmark& -& Graph \\

CVDN~\citeyearpar{thomason2020cvdn} & Materport3D & - & Low & 0 & -& -& \fullcheckmark& -& Graph \\

VLN-CE~\citeyearpar{krantz2020beyond} & Habitat, Materport3D & Specialized & Medium & 0 & \fullcheckmark& -& -& -& Discrete \\

Robo-VLN~\citeyearpar{irshad2021robovln} & Habitat, Materport3D & Specialized & Medium & 0 & \fullcheckmark& -& -& -& Continuous \\

TouchDown~\citeyearpar{chen2019touchdown} & Google Street View & - & Low & 1 & \fullcheckmark& -& -& -& Graph \\

ALFRED~\citeyearpar{shridhar2020alfred} & AI2-THOR & Specialized & Low & 120 & \fullcheckmark& -& -& -& Discrete \\

TEACh~\citeyearpar{padmakumar2022teach} & AI2-THOR & Specialized & Low & 0 & -& -& \fullcheckmark& -& Discrete \\

DialFRED~\citeyearpar{gao2022dialfred} & AI2-THOR & Specialized & Low & 0 & -& -& \fullcheckmark& -& Discrete \\

AerialVLN~\citeyearpar{liu2023aerialvln} & AirSim & Specialized & High & 25 & \fullcheckmark& -& -& -& Discrete \\

LH-VLN~\citeyearpar{song2025lhvln} & Habitat & Specialized & Medium & 216 & -& -& -& \fullcheckmark& Graph \\

GSA-R2R~\citeyearpar{hong2025general} & Habitat, Materport3D & Specialized & Medium & 150 & \fullcheckmark& -& -& -& Graph \\

VLN-PE~\citeyearpar{wang2025vlnpe} & Isaac & Full & High & 1 & \fullcheckmark& -& -& -& Hybrid \\

\rowcolor{Cerulean!10}
VLNVerse & Isaac & Full & High & 263 & \fullcheckmark& \fullcheckmark& \fullcheckmark& \fullcheckmark& Hybrid \\
\bottomrule
    \end{tabular}
    }
    \label{tab:dataset}
\end{table*}
 
The evolution of Vision Language Navigation (VLN) benchmarks has been intrinsically tied to, and constrained by, the capabilities of their underlying simulators. \Cref{tab:dataset} presents a comprehensive comparison of existing VLN datasets, dissecting their simulator capabilities, environmental diversity, and task taxonomy.

\mypara{Simulator Capabilities and Limitations.}
Early, seminal work in VLN was built on platforms like Matterport3D~\citep{chang2017matterport3d}. While foundational, this platform discretizes environments into navigation graphs. As noted in~\cref{tab:dataset}, this effectively forces agents to ``jump'' between nodes rather than using low-level actions to navigate a continuous physical space. This approach, built on static 3D scans, also abstracts away physical collision and results in lower-fidelity observations due to rendering artifacts.

Subsequent platforms, most notably Habitat~\citep{savva2019habitat}, introduced physics and continuous control, enabling tasks like VLN-CE~\citep{krantz2020beyond}. However, these environments often employ simplistic physics engines where interactions are limited to basic collision, and the continuous control is not grounded in realistic robot kinematics. Conversely, interaction-centric platforms like AI2-THOR \cite{kolve2017ai2} (used in ALFRED~\citep{shridhar2020alfred}) introduce rich physical interaction but are fundamentally designed for object manipulation rather than complex navigation. These environments are often spatially constricted, limiting their utility for training robust navigators that require extensive exploration.

\mypara{Stagnation in Environmental Diversity.}
Crucially, our comparative analysis reveals a severe stagnation in environmental diversity. As indicated by the `New Scenes \#' column in~\cref{tab:dataset}, the vast majority of subsequent benchmarks (\eg, RxR, CVDN, VLN-CE) contribute zero new environments, instead relying on re-annotations of existing Matterport3D scans. In this context, we explicitly define a ``new scene'' as an environment with unique geometry and visual appearance, distinct from previously existing datasets. Mere file format conversions do not qualify. 

For instance, while VLN-PE~\citep{wang2025vlnpe} reports utilizing 101 scenes, it comprises 90 Matterport3D scenes~\citep{chang2017matterport3d}, 10 from GRScene~\citep{wang2024grutopia}, and 1 custom scan. Although the 90 Matterport3D scenes were converted to Universal Scene Description (USD)~\citep{usd} format for Isaac Sim compatibility, they remain topologically identical to the original dataset and thus do not contribute to environmental diversity. This continued reliance on a static set of scenes risks overfitting, where agents memorize specific floor plans rather than learning generalizable navigation policies.

\mypara{Task Fragmentation.}
Parallel to simulator limitations and data stagnation, the VLN task landscape itself has become \textbf{highly fragmented}~\citep{anderson2018r2r, qi2020reverie, anderson2020rxr, thomason2020cvdn, jain2019stay, zhu2021soon, krantz2022iterative, song2025lhvln, nguyen2019help, nguyen2019vision, wang2025vlnpe, zhu2020babywalk, hong2025general, hong2020sub, chen2019touchdown}. Benchmarks are often over-specialized, isolating single instruction styles: fine-grained navigation (\eg, R2R~\citep{anderson2018r2r}, VLN-CE \cite{krantz2020beyond}, RxR~\citep{anderson2020rxr}, FGR2R~\citep{hong2020sub}); coarse-grained, goal-oriented navigation (\eg, REVERIE~\citep{qi2020reverie}, SOON~\citep{zhu2021soon}); long-horizon tasks requiring multi-stage reasoning (\eg, R4R~\citep{jain2019stay,zhu2020babywalk}, IR2R~\citep{krantz2022iterative}, LH-VLN~\citep{song2025lhvln}); and interactive, dialogue-based navigation (\eg, HANNA~\citep{nguyen2019help}, VNLA~\citep{nguyen2019vision}, CVDN~\citep{thomason2020cvdn}). This separation makes it difficult to develop and evaluate general-purpose locomotion agents.

\mypara{Advancements with VLNVerse.}
We note that while VLNVerse is built on NVIDIA's Isaac Sim, other recent benchmarks leveraging this simulator, such as BEHAVIOR-1K~\citep{li2024behavior}, Arnold~\citep{gong2023arnold}, and GRUtopia~\citep{wang2024grutopia}, have primarily focused on other embodied tasks like robotic manipulation. In contrast, VLNVerse is the first to harness the high-fidelity physics and photorealism of Isaac Sim specifically to address the core challenges of full-stack embodied navigation.

VLNVerse addresses the systemic limitations of previous works by leveraging the Isaac simulator to deliver high-fidelity, fully physical simulation with a `Hybrid' action space. It introduces the largest expansion of diversity with 263 unique scenes and is the first benchmark to unify all taxonomies (Fine-grained, Coarse-grained, Interactive, and Long-horizon), establishing a holistic standard for embodied AI research.
 \section{The VLNVerse}
The VLNVerse is designed as a decoupled three-layer architecture, as illustrated in~\cref{fig:intro}. The simulator layer serves as the foundation of the framework, encapsulating embodiment, control logic, and perception APIs. The environment layer provides high-fidelity and interactive 3D environment assets, as well as its meta-information. The task \& dataset layer builds upon the first two layers, encompassing navigation tasks, data generation pipeline, and evaluation protocol.

\begin{figure}[t!]
    \centering
    \includegraphics[width=1\linewidth]{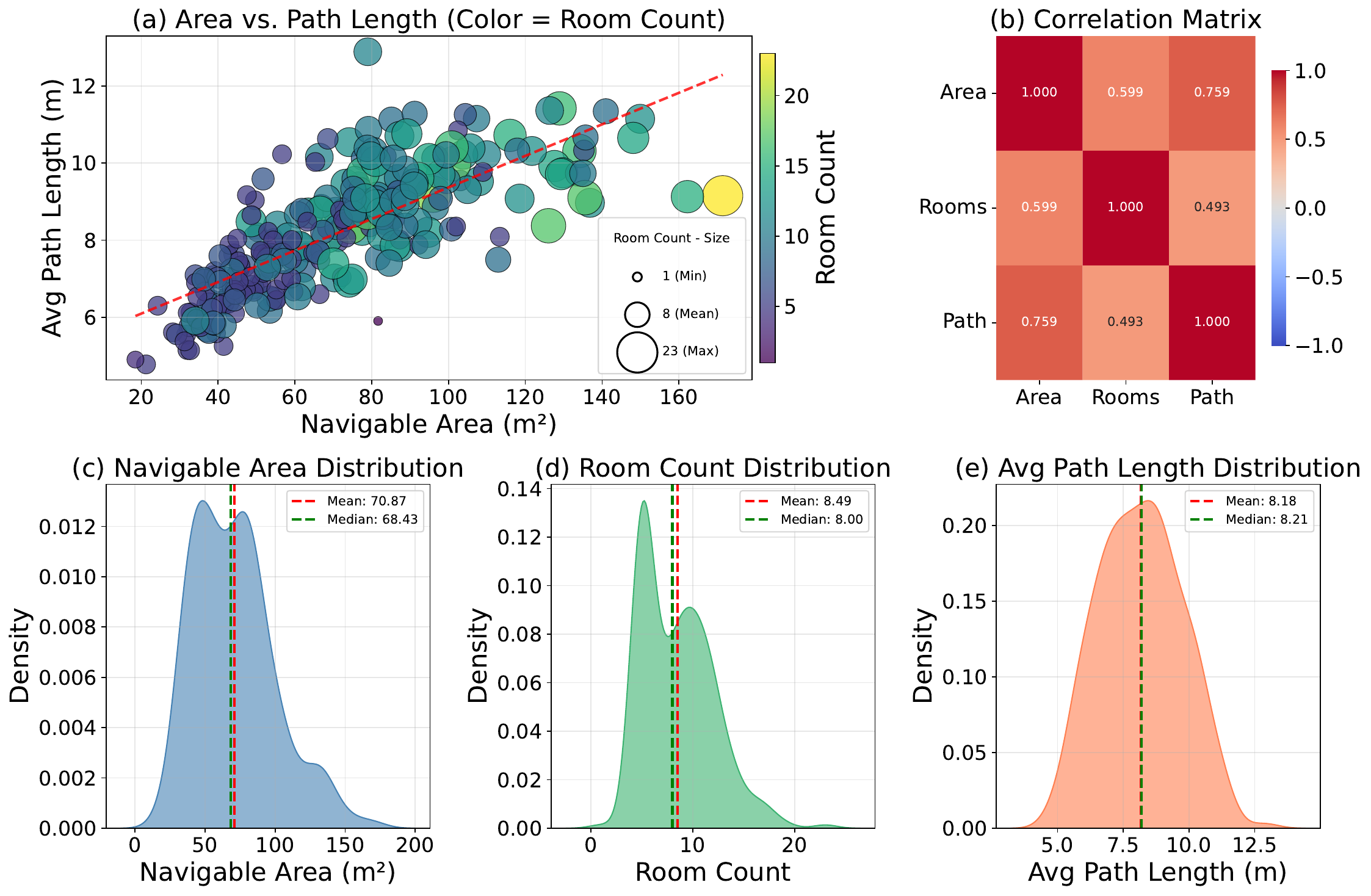}
    \caption{Distribution of navigable area, room and trajectory length.}
    \label{fig:nav_area}
\end{figure}

\subsection{Simulation Layer}
\label{sec:sim}
Our framework is designed as a full-stack solution, with the Simulator Layer serving as its foundation. This layer is built upon NVIDIA Isaac Sim to leverage its high-fidelity rendering and robust physics engine. However, our primary contribution is not the engine itself, but a novel, high-level abstraction API stack designed specifically for VLN research. This API provides: 
\textbf{(1)} \textbf{seamless integration} of Isaac Sim's physics capabilities into VLN-specific workflows; 
\textbf{(2)} \textbf{standardized interfaces} that abstract complex robot control while preserving physical realism; 
and \textbf{(3)} \textbf{universal compatibility} with various robot morphologies and sensors. 

This design eliminates the need for researchers to handle low-level robotics details, allowing them to focus on high-level navigation and interaction logic. Our Simulator Layer's design manifests these principles in three key areas: embodiment, control, and perception.

\mypara{Parameterizable Agent Embodiment.}
Instead of relying on abstract floating cameras, VLNVerse defines agents as concrete, physics-aware entities. We provide a highly parameterizable agent definition, where researchers can specify the agent's physical footprint (\eg, as a cylinder with configurable height and diameter) and kinematic properties. This ensures that all navigation is subject to realistic physical constraints, such as collision and momentum, which is critical for sim-to-real transfer.

\mypara{Physics-Aware Control and Locomotion.}
Our framework provides a unified controller interface that governs both agent locomotion and camera articulation. This interface executes actions via a physics-aware controller, simulating realistic movement rather than discrete, grid-world teleports like MP3D~\citep{chang2017matterport3d}. This abstracts the complexity of continuous control, allowing researchers to develop policies without needing to implement the underlying physics simulation.

\mypara{Configurable and Modular Perception Rig.}
To support the diverse sensory needs of modern VLN models, agents are equipped with a modular sensor rig. This API allows researchers to easily attach, detach, and configure multiple sensors (\eg, RGB, Depth, LiDAR). Crucially, all sensor parameters are exposed to the user, including sampling frequency, image resolution, and Field of View (FoV). This flexibility enables researchers to simulate different hardware trade-offs and customize the observation data pipeline to their specific research questions.

\subsection{Environment Layer}
\label{sec:assets}
\mypara{High-Fidelity Physics-Aware Environment Assets.} As the foundation of our data pipeline, this work introduces a set of 263 large-scale, diverse, and interactive 3D environments. Unlike static 3D scans, these environments are all hand-crafted as Universal Scene Description (USD)~\citep{usd} assets, where every object is fully interactive and physics-aware. For example, mirrors in our scenes realistically reflect light and other objects, and collisions are not simple binary events. In other words, an agent colliding with an obstacle will be physically deflected based on its mass and velocity. We leverage the capabilities of NVIDIA Isaac Sim to initialize these scenes, assign physical properties (\eg, mass, friction, reflectivity) to all interactable objects, and ensure a physically-grounded simulation environment.

\mypara{Meta Information and Scene Prior.} For each 3D environment, the objects are initialized with their physics and collision properties. Based on this, we generate a 2D occupancy map, which is sufficient for a general-purpose agent and serves as the navigation ground. Concurrently, we build a detailed spatial-semantic scene-graph. Since all objects (\eg, cup, chair) are trackable assets with precise coordinates and semantic labels, this scene-graph captures the spatial state and semantic relationships of all entities (\eg, cup $\rightarrow$ on $\rightarrow$ table). This graph serves as a critical ground-truth prior for subsequent path and instruction generation.

\subsection{ Task \& Dataset Layer}
\label{sec:data_pipeline}

\subsubsection{Task Definition}
\label{sec:task}
Our pipeline generates data for a unified task framework designed to cover the key axes of VLN research~\citep{anderson2018r2r,qi2020reverie,zhu2021soon,thomason2020cvdn,song2025lhvln,anderson2020rxr}:
\begin{itemize}
\item \textbf{Instruction Granularity}: We provide (1) Fine-grained Navigation (\ie, R2R-style)~\citep{anderson2018r2r}, requiring step-by-step instruction following, and (2) Coarse-grained Navigation (goal-oriented)~\citep{qi2020reverie,zhu2021soon}, where the agent must reach a target described by its properties, testing high-level understanding.
    
    \item \textbf{Input Modality}: We include (3) Visual-Reference Navigation, which extends the input modality beyond language by providing a visual cue (\eg, a photo of an object~\citep{zhu2017target}) that the agent must locate.
    
    \item \textbf{Planning Horizon}: We introduce (4) Long-Horizon Navigation, which links multiple instructions into a complex multi-stage task. This challenges long-term planning, serving as a proxy to lifelong learning scenarios.
    
    \item \textbf{Interactivity}: We support (5) Dialogue-based Navigation, enabling two-way interactive navigation where the agent can actively seek clarification from an Oracle to resolve ambiguity.
\end{itemize}

\begin{figure}[h]
    \centering
    \begin{subfigure}[b]{0.33\linewidth}
        \centering
        \includegraphics[width=0.9\linewidth]{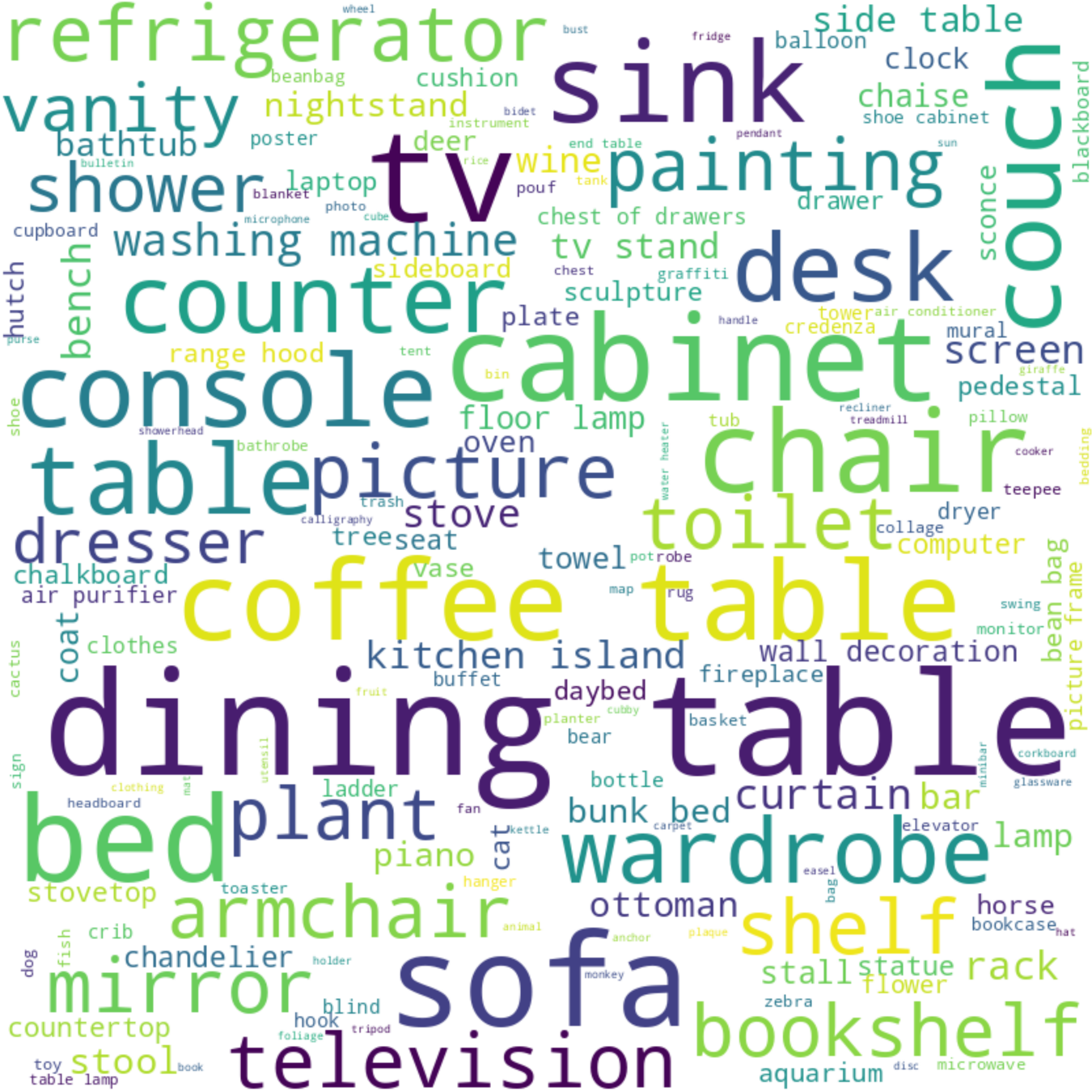}
        \caption{Landmarks.}
        \label{fig:landmark}
    \end{subfigure}
    \hfill
    \begin{subfigure}[b]{0.65\linewidth}
        \centering
        \includegraphics[width=0.9\linewidth]{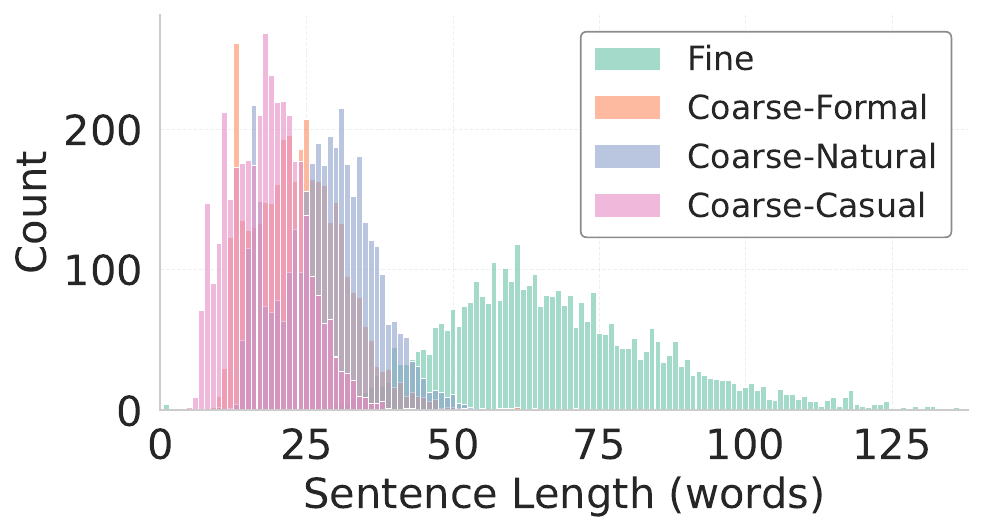}
        \caption{Instruction length distribution.}
        \label{fig:inst_length}
    \end{subfigure}

    \caption{Statistics of instruction landmarks and length.}

    \label{fig:example}
\end{figure}

\subsubsection{Scalable Generation Pipeline}
\label{sec:generation_pipeline}
To create diverse and large-scale training instances, we designed a two-stage generation pipeline:

\mypara{Physics-Aware Path Sampling.}
First, we sample navigation paths. By sampling start and goal points on the 2D occupancy map, we use an A$^*$ algorithm to find a collision-free trajectory. Critically, to ensure navigational safety and account for true embodiment, we dilate the occupancy map based on the agent's physical dimensions (\ie, its height and diameter). This ensures the sampled path is navigable by the physical agent and prevents collisions.

\mypara{Ground-Truth Instruction Generation.}
Then, we generate the ground-truth language instruction matched with the sampled path. This is a robust three-stage process:
\begin{itemize}
    \item Prior-based Initialization: We use the spatial-semantic prior (scene graph) to initialize a factually-grounded description. This grounds the instruction in semantic truth (\eg, staring at the bedroom) or spatial relationship (\eg, the mirror on the sink).
    
    \item Collaborative AI Generation: We employ a multi-agent AI system for refinement. \textbf{Describer} (Agent 1) analyzes visual primitives (\eg, video clips) from the path to describe the environment. \textbf{Verifier} (Agent 2) cross-checks the Describer's output against the scene-graph prior to identify discrepancies or hallucinations. \textbf{Synthesizer} (Agent 3) receives the verified information, task definition (\eg, fine-grained \vs coarse-grained), and linguistic style (\eg, formal \vs casual) to compose the high-quality instruction. More details are in Appendix~\ref{supp:vis}.

    \item Human Verification: As a final quality-control step, all generated instructions are presented to human volunteers. These annotators score the instructions for clarity, naturalness, and accuracy, ensuring the quality of VLNVerse.
\end{itemize}

\subsection{Evaluation Protocol and Metrics}
\label{sec:evaluation}
VLNVerse introduces a comprehensive suite of metrics that augments standard benchmarks with measures of physical robustness.

\mypara{Traditional VLN Metrics.} We evaluate agents using established metrics for task completion and efficiency. These include: Success Rate (SR), Oracle Success Rate (OSR), Success weighted by Path Length (SPL), Navigation Error (NE), Trajectory Length (TL), and normalized Dynamic Time Warping (nDTW)~\citep{ilharco2019ndtw}. Detailed definitions for each of these metrics can be found in the Appendix~\ref{supp:metric}.

\mypara{Long-Horizon Task Metrics.}
For long-horizon tasks with sequential goals, we introduce SR$_n$ to indicate the success rate for reaching the $n$-th goal (\eg, SR$_1$ for the first goal). Subsequent metrics like SR$_2$ are conditional, measuring success in reaching the second goal given that the agent successfully reached the first. This pattern continues for all intermediate goals. Finally, SR$_{\text{All}}$ measures the agent's success in stopping within the threshold distance of the \textit{final} goal, representing overall task completion.

\mypara{Physics-Aware Interaction Metrics.}
To capture the crucial aspects of embodied navigation, we introduce a new metric \textit{\textbf{Collision Rate}} (CR).
This metric quantifies the frequency of collision events the agent experiences during an episode. It directly reflects the agent's low-level obstacle avoidance capability. A low CR indicates the agent is navigating safely and smoothly.

\begin{table}[]
    \centering
        \caption{Data volume of VLNVerse. The number of episodes (instruction-trajectory pairs) are listed, with the numbers in brackets indicating the scene amounts.}

    \resizebox{0.7\textwidth}{!}
    {
    \begin{tabular}{@{}l|cccc}
    \toprule
         Taskonomy   & Train &Seen val &Unseen val &Test \\
                  \midrule
         Fine-grained   & 3963(177)  & 423(157) & 825(33) & 1325(53)   \\

         Coarse-grained & 11895(177)  & 1269(162) & 2505(33) & 3975(53)   \\
         Visual reference   & 11895(177)  & 1269(162) & 2505(33) & 3975(53)   \\
         Long horizon   & 11946(177)   & 1329(177) & 2475(33) & 3975(53)   \\
         Dialogue    & 11895(177)  & 1269(162) & 2505(33) & 3975(53) \\
    \bottomrule
    \end{tabular}}

    \label{tab:data_volume}
\end{table}
 
\subsection{Data Volume \& Statistics}
\mypara{Dataset Scale.}
The scale of the VLNVerse dataset is detailed in~\cref{tab:data_volume}. Our benchmark is built upon a total of 263 unique 3D scenes, which are exclusively split into training (177 scenes), unseen validation (33 scenes), and test (53 scenes) sets. We generate the dataset based on~\cref{sec:task}.

Specifically, for coarse-grained navigation, we generate three distinct styles (formal, natural, and casual) of instruction for each trajectory to enhance linguistic diversity. The visual reference and dialogue tasks use the identical trajectories and instructions as the coarse-grained dataset. As an auxiliary signal, we provide a reference image sampled at the final goal location for visual reference task.
For dialogue navigation, we implement an LLM-based oracle that the agent can query during navigation to gain additional information about the environment. Finally, the long-horizon tasks are constructed by randomly sampling and sequencing 2-3 sub-tasks to challenge the agent's long-term planning and memory. More details are in Appendix~\ref{supp:vis}.

\begin{wrapfigure}{r}{0.4\textwidth}
    \centering
    \vspace{-15pt}
    \includegraphics[width=\linewidth]{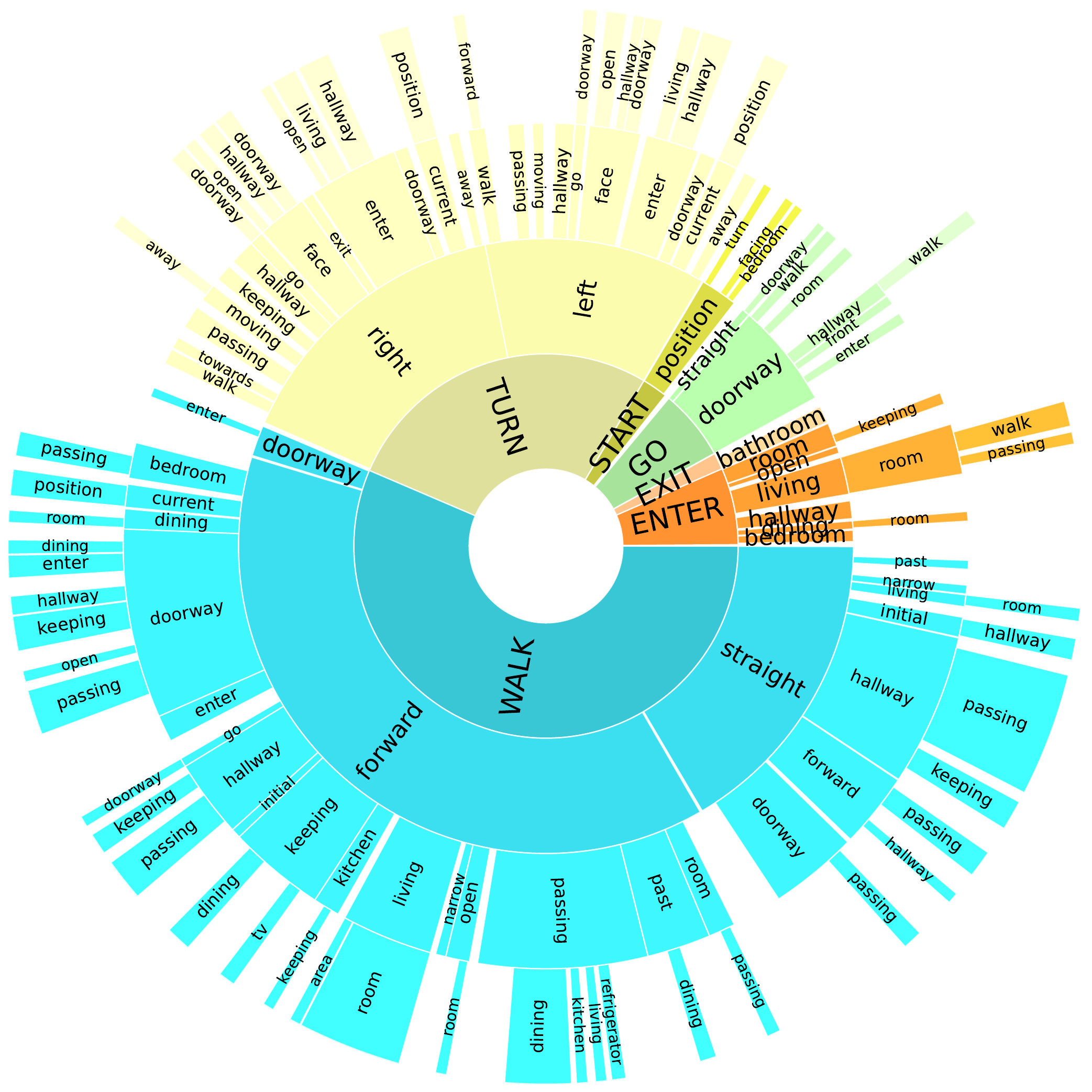}
    \caption{Hierarchy of fine-grained instruction. The white areas denote words with negligible contributions.}
    \label{fig:inst_hier}
    \vspace{-10pt} 
\end{wrapfigure}
\mypara{Instruction Statistics.} We provide a statistical analysis of the instructions in VLNVerse. As shown in~\cref{fig:inst_hier} and~\cref{fig:landmark}, the vocabularies for both navigation actions and environmental landmarks are rich, reflecting the diversity of our tasks and scenes. The~\cref{fig:inst_length} details the instruction lengths, where we observe clear patterns: Fine-grained (Fine) instructions are significantly longer, while the three styles of coarse-grained instructions (Coarse-Formal, Coarse-Natural, Coarse-Casual) are more concise and vary in length. This linguistic variety ensures our benchmark provides comprehensive coverage.

\mypara{Navigable Environment Statistics.} 
We also analyze the physical statistics of our 3D environments, as shown in~\cref{fig:nav_area}. The distributions for navigable area, room count, and path length are all highly varied, covering a wide range of scenarios rather than a simple uniform spread.
The scatter plot and correlation matrix (\cref{fig:nav_area} a\&b) confirm the intuitive positive relationships between these factors: larger areas and scenes with more rooms tend to have longer navigation paths. 
This broad and varied distribution of scene sizes and complexities ensures that VLNVerse presents a significant and diverse challenge for navigation agents.

 \begin{figure}
    \centering
    \includegraphics[width=1.0\linewidth]{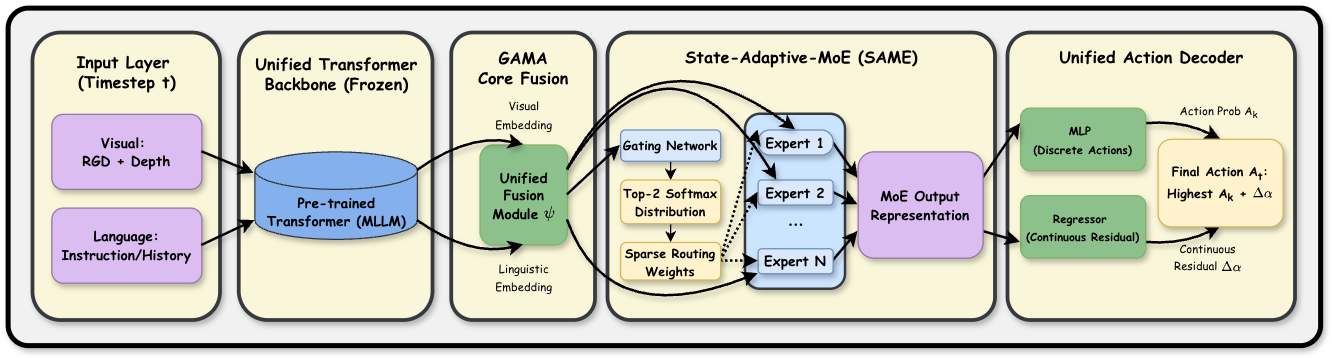}
    \caption{Framework of the proposed General-purpose Agent for Multi-task Navigation (GAMA).}
    \label{fig:gama}
\end{figure}
\section{GAMA: General-purpose Agent for Multi-task Navigation}
\label{sec:model}
To address the task fragmentation mentioned in~\cref{sec:intro}, we propose the General-purpose Agent for Multi-task Navigation, dubbed as GAMA. 
GAMA is a general-purpose framework to handle diverse observation (\ie, input views, FoV) and action spaces (discrete \vs continuous), as shown in~\cref{fig:gama}.

\subsection{Architectural Overview}
\label{sec:model_arch_philosophy}

\mypara{Unified Transformer Backbone.} GAMA's backbone leverages the pre-trained vision-language models~\citep{tan2019lxmert, wang2023scaling, li2024llama} to harness their cross-modal aligned embedding and spatio-temporal understanding for sequential tasks, allowing the robust generalization across diverse navigation scenarios.

A unified (and trainable) fusion module, $\Psi$, then serves as the GAMA-specific core. It integrates representations from two primary data streams at timestep $t$:

\begin{itemize}
    \item Visual Stream ($V_t$): RGB frames and Depth maps are processed by the frozen transformer encoder. These are concatenated and interpolated if necessary, yielding the visual embeddings.
    
    \item Language Stream ($L_t$): The instruction $L$ (\eg, single instruction or dialogue history) is tokenized and encoded by the same frozen MLLM to produce linguistic embeddings.
    
\end{itemize}

Subsequently, $\Psi$ fuses the visual and linguistic embeddings to generate a holistic state representation for the subsequent decoder.

\mypara{Unified Action Decoder.}
\label{sec:action}
Instead of separate action heads, we treat discrete actions as Dirac $\delta$ priors in a continuous space: $p(a|\psi(V_t,L_t)) = \sum_k \alpha_k \cdot \delta(a - a_k)$ where $a_k$ are 4 action primitives including stop and $\alpha_k$ are the corresponding probabilities, computed by an MLP. For continuous control, the model regresses a residual $\Delta a$ added to the most likely primitive $a_k$.

\subsection{State-Adaptive MoE (SAME)}
\label{sec:moe}

GAMA integrates the \textit{State-Adaptive Mixture-of-Experts} (SAME)~\citep{zhou2024same}, a routing mechanism designed for sequential decision-making in embodied navigation. Navigation tasks typically involve multi-step rollouts in which the relevance of linguistic cues and visual observations varies across timesteps. Conventional token-wise or task-wise MoE formulations do not align well with this temporal structure and often lead to unstable behavior or task interference~\citep{zhou2024same}. SAME instead performs routing along the \emph{time dimension}, enabling expert selection to follow the agent's state progression.

At each new timestep in which the agent enters a different location, SAME updates the selected experts based on the fused multimodal representation extracted by the policy backbone. The gating network outputs two independent top-2 softmax distributions, producing a sparse combination of experts conditioned on the current observation and instruction. By operating at the state level rather than the token level, SAME aligns the routing mechanism with the sequential nature of navigation while keeping the computation lightweight.

This formulation offers two primary benefits:
\begin{enumerate}
    \item \textbf{Compositional Multi-task Behavior.} 
    Experts learn complementary navigation behaviors, and the agent dynamically switches among them within an episode. This reduces cross-task interference and enables effective composition of diverse navigation skills.

    \item \textbf{Computational Efficiency.}
    Routing is computed once per timestep on a shared multimodal feature, avoiding token-wise routing overhead common in sparse MoE layers. This substantially lowers computational cost during long-horizon rollouts.
\end{enumerate}

We refer to Appendix~\ref{supp:models} for more implementation details and architecture design about GAMA.

 \section{Experiments}
\label{sec:exp}

\begin{table*}[t!]
\centering
\caption{Evaluating methods including discrete low-level action (CMA, Seq2Seq), continuous action (RDP), neural implict representation (HNR) and our unified model. }

\resizebox{\textwidth}{!}{
\tablestyle{2pt}{1.05}
\definecolor{Gray}{gray}{0.94}
\begin{tabular}{lcccccc>{\columncolor{Gray}}c>{\columncolor{Gray}}ccccccc>{\columncolor{Gray}}c>{\columncolor{Gray}}c}
\toprule

\multicolumn{1}{c}{\multirow{3}{*}{Methods}}
& \multicolumn{8}{c}{\textbf{Fine-grained Task}}
& \multicolumn{8}{c}{\textbf{Coarse-grained Task}}
\\
\cmidrule(r){2-9}
\cmidrule(r){10-17}

& \multicolumn{2}{c}{Val Seen}
& \multicolumn{2}{c}{Val Unseen}
& \multicolumn{4}{c}{Test}
& \multicolumn{2}{c}{Val Seen}
& \multicolumn{2}{c}{Val Unseen}
& \multicolumn{4}{c}{Test}
\\

\cmidrule(r){2-3}
\cmidrule(r){4-5}
\cmidrule(r){6-9}
\cmidrule(r){10-11}
\cmidrule(r){12-13}
\cmidrule(r){14-17}

& SR$\uparrow$ & SPL$\uparrow$
& SR$\uparrow$ & SPL$\uparrow$
& TL & NE$\downarrow$ & SR$\uparrow$ & SPL$\uparrow$
& SR$\uparrow$ & SPL$\uparrow$
& SR$\uparrow$ & SPL$\uparrow$
& TL & NE$\downarrow$ & SR$\uparrow$ & SPL$\uparrow$
\\
\midrule
\midrule

Human
& - & -
& - & -
& 12.13 & 1.74 & 86.00 & 68.60
& - & -
& - & -
& 10.23 & 1.97 & 77.20 & 62.50
\\

\midrule

CMA~\citeyearpar{wang2019reinforced}
& 37.35 & 33.36
& 31.15 & 27.92
& 8.53 & 5.24 & 28.83 & 25.54
& 32.15 & 29.06
& 35.52 & 32.48
& 7.53 & 4.72 & 31.85 & 28.67
\\

RDP~\citeyearpar{wang2025vlnpe}
& 47.28 & 41.69
& 48.60  & 42.72
& 8.18 & 4.60 & 36.38 & 32.29
& 41.61 & 37.53
& 40.85 & 35.13
& 8.35 & 4.65 & 36.60 & 32.21
\\

Seq2Seq~\citeyearpar{sutskever2014sequence}
& 32.62 & 30.39
& 35.03 & 33.37
& 5.63 & 4.79 & 30.19 & 28.15
& 31.91 & 29.68
& 33.09 & 31.00
& 6.32 & 5.00 & 27.17 & 25.53
\\

HNR~\citeyearpar{wang2024hnr}
& 36.34 & 32.10  
& 32.95 & 29.56  
& 7.98 & 4.93 & 32.24 & 29.03 
& 36.02 & 33.67  
& 40.92 & 37.27  
& 7.95 & 4.62 & 34.59 & 30.45 
\\

Ours
& 38.25 & 34.84
& 38.69 & 34.38
& 5.53 & 4.52 & 37.72 & 33.85
& 42.45 & 38.89
& 48.30 & 41.50
& 3.58 & 4.23 & 37.20 & 31.02
\\

\bottomrule
\end{tabular}}
\label{tab:baseline}
\end{table*}
 
\subsection{Experimental Setup}

\mypara{Baselines.} To provide a rigorous benchmark, we conduct a comprehensive evaluation of existing methods, which we group into two primary categories:
\begin{itemize}
    \item Classic \& Specialized Models: This group includes specialized VLN models: discrete low-level action (Seq2seq~\citep{sutskever2014sequence} and CMA~\citep{wang2019reinforced}), continuous action (RDP~\citep{wang2025vlnpe}), and neural implicit representation method HNR~\citep{wang2024hnr}.

    \item Foundation Model: We also evaluate InternNav-N1~\citep{internnav2025}, NFM~\citep{zhang2025nfm}, and UniNavid~\citep{zhang2024uni} in zero-shot setting.
    \item Large Model Agents: This group tests the generalization of modern large models. We implement two variants with text-summary history~\citep{zhou2024navgpt,zhou2025navgpt2} and text-map history~\citep{chen2024mapgpt} in QwenVL3-4B~\citep{yang2025qwen3}.
\end{itemize}

\mypara{Implementation Details.} For a fair comparison, all baseline models are trained strictly in an offline training regime using their default, out-of-the-box experimental settings.

\mypara{Training Paradigm.} We test two kinds of training paradigms: 1) Offline Training (\cref{sec:offline}): The navigation model is trained to mimic expert actions using the static data; 2) Online Fine-tuning (\cref{sec:online}): To adapt to the dynamic simulator, we fine-tune the model using online training. The agent is allowed to explore and interact with the environment, forcing it to learn a robust physics-aware policy.

\definecolor{tableheadgray}{gray}{0.9}

\begin{table}[htb]
\centering
\caption{Performance Comparison of Navigation Foundation Model on test split.}
\resizebox{0.6\textwidth}{!}{
\begin{tabular}{lcccccc}
\toprule
Model&TL&NE$\downarrow$&SR$\uparrow$&OSR$\uparrow$&SPL$\uparrow$&CR$\downarrow$\\
\midrule
\midrule

\rowcolor{tableheadgray}\multicolumn{7}{l}{\emph{Fine-grained Task}}\\
InternNav-N1~\citeyearpar{internnav2025}&9.23&5.68&28.95&38.69&25.00&-\\
NFM~\citeyearpar{zhang2025nfm}&8.58&4.13&51.59&69.68&32.4&19.35\\
UniNavid~\citeyearpar{zhang2024uni}&12.35&4.11&45.74&67.02&26.91&12.04\\

\midrule

\rowcolor{tableheadgray}\multicolumn{7}{l}{\emph{Coarse-grained Task}}\\
InternNav-N1~\citeyearpar{internnav2025}&4.00&5.71&17.51&23.32&16.54&-\\
NFM~\citeyearpar{zhang2025nfm}&6.52&4.93&38.02&45.93&23.15&24.22\\
UniNavid~\citeyearpar{zhang2024uni}&10.33&5.05&29.68&42.70&13.47&15.59\\

\bottomrule
\end{tabular}
}
\label{tab:foundation}
\end{table}
 
\subsection{Baseline Performance on VLNVerse}
\label{sec:offline}
We benchmark several methods on VLNVerse using offline training. We pre-render all observations on the trajectory. All experiments are implemented using the PyTorch framework and conducted on NVIDIA RTX 4090 GPUs. 
The CMA and Seq2Seq models are trained on a single GPU with a batch size of 2, requiring approximately one day to reach convergence.  Conversely, the RDP, HNR, and GAMA models are trained across 4 GPUs utilizing \texttt{DataParallel}, with a total effective batch size of 4, requiring approximately two days. We optimize all models using AdamW with a learning rate of $1 \times 10^{-4}$. The maximum trajectory length is capped at 200 steps. 

The results summarized in~\cref{tab:baseline} highlight the significant challenge posed by the VLNVerse benchmark. On the fine-grained test split, our unified model demonstrates the strongest performance among the evaluated methods, achieving the highest SR at 37.72\% and the highest SPL at 33.85\%. This is closely followed by the diffusion-based model, RDP~\citep{internnav2025}, which also shows robust results with an SR of 36.38\% and an SPL of 32.29\%. In contrast, other methods show lower performance. The neural implicit representation model, HNR~\citep{wang2024hnr}, achieves an SR of 32.24\%, while classic discrete-action methods like Seq2Seq~\citep{sutskever2014sequence} (30.19\% SR) and CMA~\citep{wang2019reinforced} (28.83\% SR) lag further behind.

We also evaluate large-scale, pre-trained navigation foundation models in~\cref{tab:foundation}. It is noteworthy that while these models have demonstrated strong capabilities on established discrete-action benchmarks~\citep{anderson2018r2r,savva2019habitat}, their performance tends to fluctuate when migrated to our more complex, physics-aware continuous environment. Among these methods, NFM~\citep{zhang2025nfm} achieves the highest success rates at 51.59\% and 38.02\% for fine-grained and coarse-grained tasks, respectively. UniNavid~\citep{zhang2024uni} posts comparable success rates of 45.74\% and 29.68\%, while InternNav-N1~\citep{internnav2025} shows the lowest performance in this group with success rates of 28.95\% and 17.51\%.

The overall performance gap suggests our environment is challenging. Traditional discrete-action methods (CMA, Seq2Seq) appear to struggle. Models better suited for continuous control, such as the diffusion-based RDP and our unified model, achieve higher success. Furthermore, the significant drop in SPL for foundation models like InternNav-N1 and UniNavid indicates they struggle with path efficiency in our setting, even when they successfully find the target.

\begin{table}[t]
    \centering
    \caption{Performance comparison of VLN agents with Tel-Hop and strict collision detection mechanisms.}
    \label{tab:zero_shot_main}
    \resizebox{1\columnwidth}{!}{
    \tablestyle{2pt}{1.05}
    \begin{tabular}{l l r r r r r r r r r r r r r r} 
        \toprule
        \multirow{3}{*}{\textbf{Agent}} & \multirow{3}{*}{\textbf{Collision}} & \multicolumn{7}{c}{\textbf{Fine-grained Task}} & \multicolumn{7}{c}{\textbf{Coarse-grained Task}} \\
        \cmidrule(lr){3-9} \cmidrule(lr){10-16} 
        & & TL & NE$\downarrow$ & SR$\uparrow$ & OSR$\uparrow$ & SPL$\uparrow$ & nDTW$\uparrow$ & CR$\downarrow$ & TL & NE$\downarrow$ & SR$\uparrow$ & OSR$\uparrow$ & SPL$\uparrow$ & nDTW$\uparrow$ & CR$\downarrow$ \\
        \midrule
        \midrule
        
        \multirow{2}{*}{Text-history~\citeyearpar{zhou2024navgpt}} & Tel-Hop & 24.22 & 6.10 & 19.27 & 61.46 & 8.13 & 52.61 & 32.13 & 22.36 & 5.24 & 31.25 & 55.73 & 7.25 & 49.60 & 28.99 \\
         & Strict & 6.92 & 6.10 & 19.27 & 21.88 & 14.93 & 42.66 & 56.93 & 6.57 & 5.75 & 23.96 & 28.65 & 15.34 & 43.97 & 52.31 \\
        \midrule

        \multirow{2}{*}{Map-history~\citeyearpar{chen2024mapgpt}} & Tel-Hop & 34.23 & 5.62 & 25.53 & 55.32 & 7.57 & 43.38 & 27.56 & 52.68 & 4.33 & 49.48 & 69.79 & 12.32 & 46.16 & 36.16 \\
         & Strict & 5.43 & 6.16 & 16.67 & 23.44 & 13.23 & 42.85 & 48.20 & 7.05 & 5.52 & 22.92 & 31.25 & 14.23 & 45.55 & 54.71 \\
        \midrule

        \multirow{2}{*}{Text-history~\citeyearpar{zhou2024navgpt}+CoT} & Tel-Hop & 21.22 & 5.25 & 28.72 & 50.53 & 9.91 & 44.12 & 20.70 & 36.48 & 4.22 & 43.75 & 70.83 & 9.60 & 47.01 & 26.62 \\
         & Strict & 5.07 & 6.23 & 17.19 & 18.75 & 14.78 & 41.96 & 43.23 & 6.33 & 5.53 & 21.35 & 29.17 & 14.14 & 45.25 & 38.58 \\
        \midrule

        \multirow{2}{*}{Map-history~\citeyearpar{chen2024mapgpt}+CoT} & Tel-Hop & 40.92 & 4.51 & 42.19 & 72.40 & 11.32 & 45.22 & 25.30 & 24.18 & 4.32 & 43.23 & 60.94 & 15.03 & 49.68 & 18.57 \\
         & Strict & 5.43 & 5.88 & 16.67 & 22.40 & 14.02 & 43.13 & 43.65 & 5.51 & 5.36 & 26.04 & 31.77 & 18.61 & 45.52 & 33.42 \\
        \bottomrule
    \end{tabular}
    }
\end{table}
 
\subsection{Zero-shot Performance on VLNVerse}
\label{sec:zeroshot}
We evaluate both text-history and map-history agents~\citep{zhou2024navgpt, long2023discuss, qiao2024open, shi2025smartway, shi2025fast, li2025boosting}, utilizing QwenVL3-4B~\citep{yang2025qwen3} as the LLM backbone. Following previous works~\citep{qiao2024open, shi2025smartway, qiao2025navbench}, we randomly sampled 200 test trajectories across all test scenes for zero-shot evaluation. We established two distinct evaluation settings: 
1) \textbf{Strict:} Navigation occurs under full physical constraints. An episode fails upon a noticeable collision, defined as an obstacle displacing the agent by 0.1m. 
2) \textbf{Tel-Hop:} The agent teleports between waypoints, bypassing obstacles during transit. Physical constraints are only applied at the destination. If the target is inside an obstacle, the physics engine resolves it by moving the agent to the nearest reachable location, and the episode proceeds. This setting mimics discrete-environment evaluations but still penalizes physically invalid final waypoints.

Separately from these settings, we also investigate the impact of Chain-of-Thought (CoT)~\citep{wei2022chain} prompting. This strategy, designed to enhance the LLM's reasoning process, is applied as an augmentation to agents in both the Strict and Tel-Hop settings.

Our findings reveal a critical gap between idealized and physics-based navigation. As seen in~\cref{tab:zero_shot_main}, agent performance is severely degraded in the Strict physics-based setting. For instance, the map-history agent's success rate drops from 25.5\% in the Tel-Hop setting to 16.7\% in the Strict setting. This performance gap stems from the fact that these MLLM agents are designed for discrete, collision-free graphs, where waypoint predictions do not account for physical embodiment. The Collision Rate statistics strongly support this: in the Strict setting, agents frequently collide, leading to immediate episode failure.

Strikingly, when we introduce CoT prompting, performance in the Tel-Hop setting improves further, yet performance in the Strict setting remains low. This result strongly confirms that physical collision is the primary bottleneck. While enhanced reasoning improves performance in an idealized, collision-relaxed setting, it is insufficient to overcome the physical-interaction challenge.

These results suggest that while the MLLM's in-context reasoning possesses sufficient zero-shot capability for high-level navigation logic, current agent models are extremely sensitive to collision-aware environments.

\definecolor{tableheadgray}{gray}{0.9}

\begin{table}[t!]
\centering
\caption{Performance comparison of VLN agents across visual reference, dialogue, and long-horizon navigation tasks.}

\label{tab:combined_tasks}
\resizebox{1\columnwidth}{!}{ 
\tablestyle{8pt}{1.0}
\begin{tabular}{l|ccccccc|ccc}
\toprule

Agent & TL & NE$\downarrow$ & SR$\uparrow$ & OSR$\uparrow$ & SPL$\uparrow$ & nDTW$\uparrow$ & CR$\downarrow$ & SR1$\uparrow$ & SR2$\uparrow$ & SR3$\uparrow$ \\
\midrule
\midrule

\rowcolor{tableheadgray}\multicolumn{11}{l}{\emph{Visual Reference}}\\
Text-history~\citeyearpar{zhou2024navgpt}& 24.43 & 5.70 & 29.26 & 55.32 & 7.79 & 47.76 & 32.26 & - & - & - \\
Map-history~\citeyearpar{chen2024mapgpt} & 54.51 & 4.85 & 42.55 & 77.13 & 5.19 & 46.29 & 36.20 & - & - & - \\
\midrule

\rowcolor{tableheadgray}\multicolumn{11}{l}{\emph{Dialogue-based}}\\
Text-history~\citeyearpar{zhou2024navgpt} & 24.57 & 2.92 & 84.57 & 88.30 & 35.20 & 60.46 & 30.92 & - & - & - \\
Map-history~\citeyearpar{chen2024mapgpt} & 25.36 & 4.14 & 67.02 & 74.47 & 30.46 & 53.62 & 30.33 & - & - & - \\
\midrule

\rowcolor{tableheadgray}\multicolumn{11}{l}{\emph{Long-horizon}}\\
Text-history~\citeyearpar{zhou2024navgpt} & 39.77 & 5.68 & 20.74 & - & 6.45 & 33.99 & 19.56 & 34.04 & 12.23 & 4.79 \\
Map-history~\citeyearpar{chen2024mapgpt} & 49.92 & 5.82 & 25.5 & - & 2.13 & 47.76 & 33.49 & 77.13 & 46.28 & 10.64 \\

\bottomrule
\end{tabular}

}
\end{table}

\subsection{Communicative and Long-horizon Navigation}
We further evaluated the zero-shot performance of the map-history agent on more complex tasks, building upon the coarse-grained navigation setup. All experiments in this section were conducted in the \textbf{Tel-Hop} setting to isolate the agent's high-level reasoning capabilities from the physical collision bottleneck identified in~\cref{sec:zeroshot}. We tested three distinct scenarios.

\paragraph{Visual Reference Navigation.}
Here, the agent received a reference image of the target area in addition to the textual instruction. As seen in~\cref{tab:combined_tasks}, this supplementary visual cue did not yield a meaningful performance improvement. The SR remained stagnant at 42.6\% (compared to 42.2\% for the map-history agent with CoT in~\cref{tab:zero_shot_main}). Interestingly, the average Trajectory Length (TL) increased from 40.9 to 54.5. This suggests the MLLM may have struggled to properly ground the visual cue, perhaps searching for a perfect visual match, which prevented it from correctly identifying the stopping condition.

\paragraph{Dialogue-Based Navigation.}
In this setting, the agent could query an oracle LLM for environmental information and navigational advice. This intervention, detailed in~\cref{tab:combined_tasks}, led to a massive performance boost, with the map-history agent's SR jumping from 42.2\% to 67.0\%.

This dialogue-based result is highly significant and leads to two key conclusions. First, the oracle LLM, which had no prior fine-tuning on our environment, could process the context and provide effective guidance. This demonstrates that the high-quality, realistic rendering and logical structure of VLNVerse are highly compatible with the reasoning capabilities of large models. Second, the ability to engage in dialogue effectively reduces agent confusion and resolves ambiguity, drastically improving the navigation success rate. This strongly supports the value of interactive agents for complex, real-world tasks.

\paragraph{Long-Horizon Navigation.}
In this task, the agent was given an instruction to navigate to 2$\sim$3 goals sequentially. As shown in~\cref{tab:combined_tasks}, the agent performed well at reaching the first goal, achieving an SR1 of 77.1\%. However, performance dropped significantly for subsequent goals, with SR2 at 46.3\% and SR3 falling to just 10.6\%. While the final goal success rate (SR-All) was 25.5\%, the sharp decline in sequential success indicates that the model struggles to maintain a complex, multi-stage plan over extended horizons.

\begin{table}[!t]
    \centering
    \caption{Performance of Online fine-tuning on VLN-BERT agent.}
    \resizebox{0.75\columnwidth}{!}{
    \begin{tabular}{lccccccc}
        \toprule
        VLN-BERT~\citeyearpar{hong2020recurrent} & TL & NE$\downarrow$ & SR$\uparrow$ & OSR$\uparrow$ &SPL$\uparrow$ & nDTW$\uparrow$ & CR$\downarrow$ \\
        \midrule
        Zero-shot & 26.2 & 6.76 & 3.4 & 42.3 & 1.5 & 31.3 & 52.9 \\
        Fine-tuned(1 epoch) & 23.7 & 6.13 & 11.1 & 53.3 & 5.7 & 33.2 & 31.5 \\
        Fine-tuned(10 epochs) & 4.28 & 5.23 & 25.7 & 29.7 & 23.7 & 49.7 & 19.5 \\
        \bottomrule
    \end{tabular}
    }
    \label{tab:online_sup}
\end{table}
 \subsection{Online Fine-Tuning}
\label{sec:online}

\begin{figure*}[t]
    \centering
    \includegraphics[width=\textwidth]{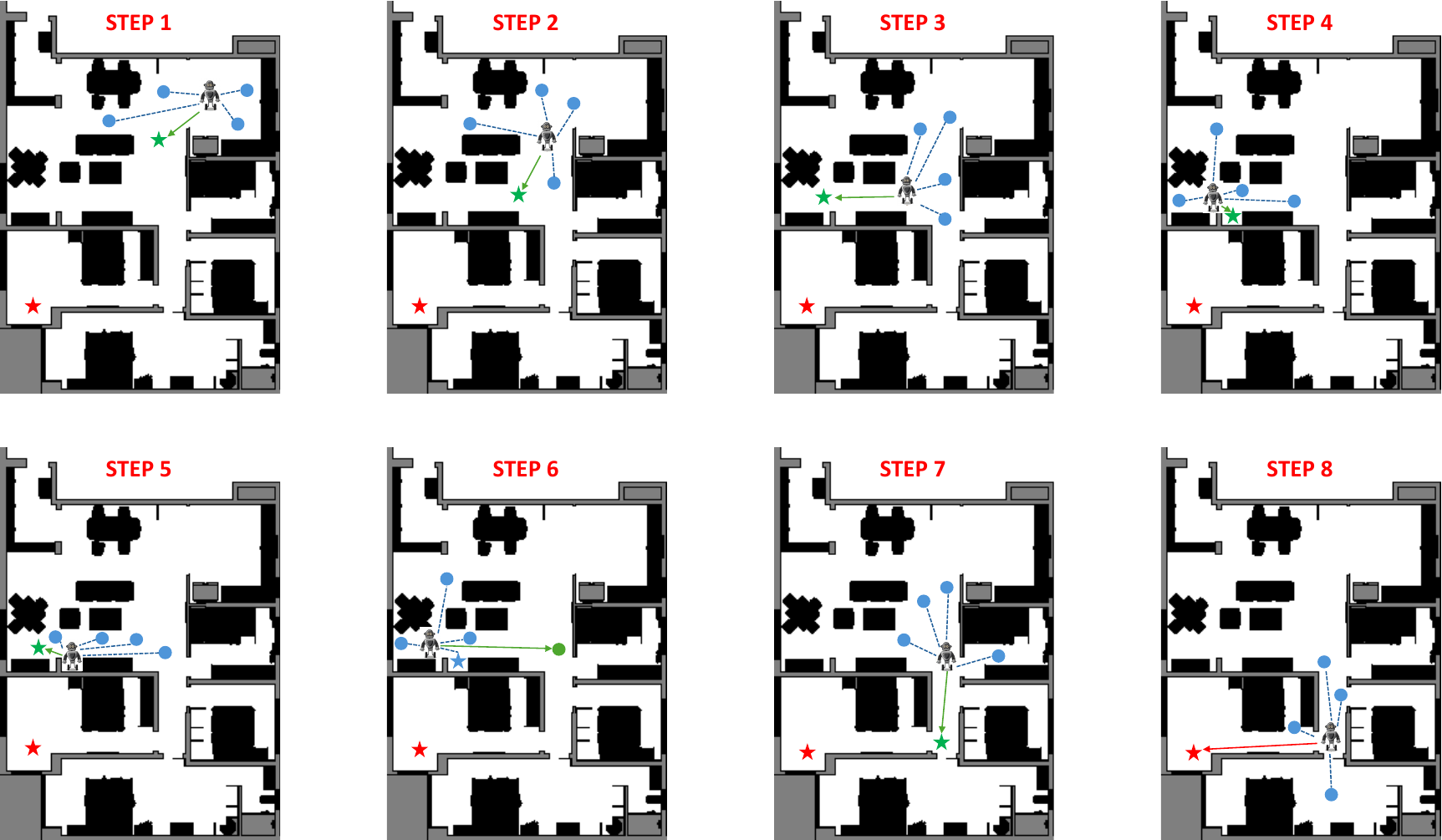}
    \caption{Visualization of trajectories and predicted waypoints.}
    \label{fig:sup_online}
    \vspace{-8pt}
\end{figure*}

\mypara{Loss Function.} Methods based on waypoint prediction~\citep{hong2022bridging, an2023etpnav, an2022bevbert, wang2023gridmm, li2025ground, wang2024sim} are particularly challenging for purely offline training, as the absence of an interactive feedback mechanism in the new environment prevents the correction of accumulated errors from waypoint prediction, leading to rapid navigation failure. A key feature of VLNVerse is its support for online fine-tuning. To demonstrate this, we investigate the effectiveness in adapting existing models pre-trained on other benchmarks. Specifically, we employ VLN-BERT~\citep{hong2022bridging} and evaluated its zero-shot performance in VLNVerse. 
The agent in our experiments is trained by waypoint-based methods~\citep{hong2022bridging}, using Imitation Learning (IL) with a cross-entropy loss on the action probabilities $p_{t}$ and the oracle action $a^{*}_{t}$. The loss is defined as:
\begin{equation}
\mathcal{L}_{IL} = - \sum_{t} a^{*}_{t}\log\left(p_{t}\right),
\end{equation}
where the oracle action $a^{*}_{t}$ corresponds to the waypoint (\tikz[baseline=-0.6ex]\node[star,star points=5,star point height=0.3em,draw=green,fill=green,inner sep=1pt] {}; in~\cref{fig:sup_online}) with the shortest geodesic distance to the target among all predicted candidates(\tikz[baseline=-0.6ex]\draw[customblue,fill=customblue] (0,0) circle (0.5ex); in~\cref{fig:sup_online}). The oracle stop is successful when the geodesic distance between the agent and the target(\tikz[baseline=-0.6ex]\node[star,star points=5,star point height=0.3em,draw=red,fill=red,inner sep=1pt] {}; in~\cref{fig:sup_online}) is less than 1.5 meters and stays in open-space. Note that in rare cases, the waypoint predictor might not be able to return an oracle waypoint that brings the agent closer to the target. In order to allow the agent to explore the environment while learning from teacher actions, we control the agent with schedule sampling~\citep{bengio2015scheduled} to probabilistically sample an action between the oracle and the prediction at each step.

\mypara{Experimental Setup.} The model used in our experiments (VLN-BERT~\citep{hong2022bridging}) is initialized from the pre-trained PREVALENT model~\citep{hao2020towards}. Additionally, the weights of our navigation policy are initialized from a model pre-trained with VLN-CE in the Habitat simulator. We select the first 30 scenes from the VLNVerse for online training. This subset selection is motivated by the computational demands inherent to online training, which involves real-time rendering and dynamic scene interactions. To ensure consistency, the validation and testing are also performed using the validation and test splits corresponding to these same 30 scenes. We train the VLN-BERT for 10 epochs with a batch size of 4, taking approximately 15 hours to complete using 4 NVIDIA RTX 3090 GPUs.

\mypara{Results.} As shown in~\cref{tab:online_sup}, due to domain shift, VLN-BERT exhibits poor performance with SR of 3.4\%. The performance of VLN-BERT increases to $11.09\%$ SR after only one epoch of brief training conducted on a subset consisting of the first 30 scenes in VLNVerse. We then train it for 10 epochs and achieves $25.7\%$ SR and $23.7\%$ SPL. This result strongly validates the design of our benchmark. It shows that the offline and online training paradigms provided by VLNVerse are complementary and effective.

\mypara{Agent-Environment Interaction Visualization.}
\cref{fig:sup_online} visualizes the agent's trajectories and the predicted waypoints at each step. As shown by the waypoints (\tikz[baseline=-0.6ex]\draw[customblue,fill=customblue] (0,0) circle (0.5ex);) in panoramas and the occupancy maps, most of the predictions are nicely positioned at accessible spaces, pointing towards explorable directions around the agent. Thanks to the predicted waypoints, the agent often only needs to make a few decisions to complete a long navigation task. However, in some cases, such as Steps 4 and 5, the waypoint predictor might not be able to return a waypoint that immediately reduces the distance to the target due to environmental constraints (\eg, the target is separated from the agent by a wall). Consequently, the agent must navigate a U-shaped path to bypass the obstacle rather than moving directly towards the target. In order to allow the agent to successfully explore such complex topologies while learning from teacher actions, we control the agent using schedule sampling~\citep{bengio2015scheduled}, it probabilistically samples an action between the oracle and the prediction at each step, enabling the agent to break out of local optima and effectively navigate around obstacles. A specific example is shown in Step 6 in~\cref{fig:sup_online}, where the agent executes a predicted candidate waypoint(\tikz[baseline=-0.6ex]\draw[green,fill=green] (0,0) circle (0.5ex);) rather than the oracle action(\tikz[baseline=-0.6ex]\node[star,star points=5,star point height=0.3em,draw=customblue,fill=customblue,inner sep=1pt] {};) determined by the geodesic distance.

\section{Conclusion}
This work introduces VLNVerse, a comprehensive, scalable, and physically-grounded benchmark for Vision-Language Navigation. Our framework is built on a decoupled three-layer architecture that separates the responsibilities of the Agent (Simulator Layer), World (Environments Layer), and Benchmark (Task \& Dataset Layer). Our primary contribution is a step towards full-stack development in embodied VLN, including (1) a new set of 263 large-scale, fully interactive 3D environments; (2) a physics-aware agent embodiment; (3) a comprehensive evaluation on existing literature; and (4) a unified MoE-based model.

\subsubsection*{Acknowledgments}

We thank Zihan Wang for the helpful discussion on the HNR implementation, and Jiazhao Zhang for the UniNavid and NFM results. We also appreciate Jian Zhou’s advice on the Isaac Simulator, and Yang Li for his assistance with GPU scheduling.

\bibliography{iclr2026_conference}
\bibliographystyle{iclr2026_conference}

\clearpage
\appendix

\section*{Appendix}
\addcontentsline{toc}{section}{Appendix}

\setcounter{section}{0}
\renewcommand{\thesection}{\Alph{section}}

\section{Model Details \& Training Recipe}
In this section, we provide comprehensive implementation details for the models and training pipelines discussed in the main paper. We first elaborate on the architecture of our unified baseline, GAMA, followed by a description of the comparative baseline methods. Finally, we present the specific experimental setups and hyperparameter configurations for both offline and online training phases.
\label{supp:models}

\subsection{GAMA: General-purpose Agent for Multi-task Navigation}
\label{sec:appendix_gama}

Our unified baseline GAMA integrates a hierarchical navigation backbone~\citep{wang2024hnr} with a novel state-adaptive navigation policy~\citep{zhou2024same}. In the following, we detail the implementation of the hierarchical representation backbone in~\cref{sec:gama_hnr_backbone}, followed by the state-adaptive navigation policy in~\cref{sec:gama_policy_combined}

\subsubsection{Hierarchical Neural Radiance Backbone}
\label{sec:gama_hnr_backbone}

Unlike standard recurrent policies that rely solely on current observations, GAMA adopts the lookahead exploration strategy from HNR~\citep{wang2024hnr}. It utilizes a hierarchical neural radiance representation to hallucinates semantic representations of future environments. The backbone consists of three key stages:

\paragraph{Feature Cloud Encoding:} 
The agent maintains a feature cloud $\mathcal{M}$ that stores fine-grained visual semantics and 3D spatial information. At step $t$, it extracts grid features $\mathbf{g}_{t,j}$ from the observation using a vision encoder (\eg, CLIP-ViT) and projects them into 3D space using depth $\mathbf{d}_{t,j}$ and camera intrinsics $\mathbf{K}$:
\begin{equation}
    P_{t,j} = \left[ \mathbf{d}_{t,j} \mathbf{R}^{-1} \mathbf{K}^{-1} [h, w, 1]^T - \mathbf{T} \right]^T,
\end{equation}
where $P_{t,j}$ is the 3D coordinate. The cloud is updated as $\mathcal{M}_{t} = \mathcal{M}_{t-1} \cup \{[\mathbf{g}_{t,j}, P_{t,j}, \theta_{t,j}, s_{t,j}]\}$, including orientation $\theta$ and scale $s$.

\paragraph{Future Prediction via Volume Rendering:} 
To predict the feature map of a candidate lookahead view, the model employs neural volume rendering. For a sampled point $\mathcal{P}_n$ along a camera ray, it retrieves the $K$-nearest features from $\mathcal{M}$ using a KD-Tree. To ensure translational invariance, relative positional embeddings are computed:
\begin{equation}
    \mathbf{q}_k = \text{LN}(\mathbf{W}_{1}[P^{rel}_{k}, \theta^{rel}_{k}, s_{k}]),
\end{equation}
where $P^{rel}$ and $\theta^{rel}$ are relative coordinates and orientation. These are aggregated to produce a density $\sigma_n$ and latent vector $\mathbf{r}_n$. The final region feature $\mathbf{R}_{h,w}$ is computed via integration:
\begin{equation}
    \mathbf{R}_{h,w} = \sum_{n=1}^{N} \tau_n (1 - \exp(-\sigma_n \Delta_n)) \mathbf{r}_n, \quad \tau_n = \exp\left(-\sum_{i=1}^{n-1} \sigma_i \Delta_i\right).
\end{equation}
These rendered features allow the agent to ``see" potential future states without physical movement.

\paragraph{Topological Graph Construction:} 
The predicted future views are organized into a topological graph containing visited nodes, candidate nodes, and lookahead nodes. The graph is encoded using Graph-Aware Self-Attention (GASA), which incorporates the geodesic distance matrix $E$ into the attention mechanism:
\begin{equation}
    \text{GASA}(\mathcal{V}) = \text{Softmax}\left( \frac{\mathcal{V}\mathbf{W}_q(\mathcal{V}\mathbf{W}_k)^T}{\sqrt{d}} + E\mathbf{W}_e \right) \mathcal{V}\mathbf{W}_v.
\end{equation}
This graph representation $\mathcal{V}$ serves as the input to the navigation policy.

\subsubsection{Multimodal Navigation Policy}
\label{sec:gama_policy_combined}

Building upon the topological representations from the backbone, GAMA employs a State-Adaptive Mixture of Experts (SAME) to execute precise navigation actions aligned with language instructions.

\paragraph{Instruction-Vision Alignment:}
At each timestep, the agent receives the visual features $\hat{v}_t$ (derived from the current view) and updates the topological map $\hat{\mathcal{G}}_t$. To align the visual context with the language instruction $\mathcal{W}$, a local cross-modal encoder excites visual features conditioned on text via Cross-Attention:
\begin{equation}
    \text{CrossAttn}(\hat{v}_t, \hat{\mathcal{W}}) = \text{Softmax}\left( \frac{\hat{v}_t W_q (\hat{\mathcal{W}} W_k)^T}{\sqrt{d}} \right) \hat{\mathcal{W}} W_v.
    \label{eq:cross_attn}
\end{equation}
The output embedding for the current view is $\hat{v}'_t$. Similarly, a global cross-modal encoder encodes the map to yield map node embeddings $\hat{v}'_{g,i}$:
\begin{equation}
    \hat{\mathcal{G}}'_t = \text{CrossAttn}(\hat{\mathcal{G}}_t, \hat{\mathcal{W}}).
\end{equation}

\paragraph{State-Adaptive Navigation Policy:}
In the previous HNR~\citep{wang2024hnr}, navigation scores are computed via static Feed-Forward Networks. GAMA replaces this with the SAME mechanism to handle task fragmentation. 
First, a multimodal routing feature $\mathbf{x}_r^{\text{multi}}$ is computed to capture the alignment between the visual context and the instruction. We utilize the global representation of the current monocular view aligned with the instruction's $[\text{CLS}]$ token:
\begin{equation}
    \mathbf{x}_r^{\text{multi}} = W_m \left[ \hat{v}_t^{\text{CLS}} \mathbin{;} \hat{\mathcal{W}}^{\text{CLS}} \right],
\end{equation}
where $W_m$ is a projection layer. The router $\mathcal{R}$ predicts activation probabilities for $N$ experts:
\begin{equation}
    \mathcal{P}(\mathbf{x}_r) = \text{Softmax}(W_r \mathbf{x}_r^{\text{multi}}).
\end{equation}
The local observation features $\hat{v}'_t$ and global map features $\hat{v}'_{g,i}$ are then processed by the top-$k$ selected experts. The navigation scores are calculated as:
\begin{equation}
    \begin{aligned}
        s^l &= \sum_{k \in \mathcal{T}} \mathcal{P}(\mathbf{x}_r)_k \cdot f_k^{\text{local}}(\hat{v}'_t), \\
        s^g_i &= \sum_{k \in \mathcal{T}} \mathcal{P}(\mathbf{x}_r)_k \cdot f_k^{\text{global}}(\hat{v}'_{g,i}).
    \end{aligned}
\end{equation}
Finally, the global and local scores are fused using a learnable parameter $\sigma_t$ to determine the action:
\begin{equation}
    s_i = \sigma_t s^l + (1 - \sigma_t)s^g_i.
\end{equation}
To ensure diverse expert usage, we apply a load balancing loss $\mathcal{L}_{\text{balance}} = N \sum \mathcal{F}_i \mathcal{D}_i$ during training.

\subsection{Baseline Models}
To evaluate the effectiveness of our approach, we compare it against a comprehensive set of baselines ranging from traditional recurrent architectures to recent foundation models. We categorize these baselines into three groups based on their action spaces and architectural paradigms.

\subsubsection{Recurrent and Graph-based Methods}

\paragraph{Sequence-to-Sequence (Seq2Seq).}
This baseline, adapted from~\citep{krantz2020beyond}, represents a foundational VLN architecture. It comprises three main modules: 
\begin{enumerate}
    \item An \textbf{instruction encoder} that processes the language input ($I = \{w_i\}_{i=1}^{L}$) using an LSTM on GLoVE embeddings~\citep{pennington-etal-2014-glove}.
    \item An \textbf{observation encoder} that uses a pre-trained ResNet50~\citep{he2016deep} for RGB visual features ($V_t$) and a separate pre-trained model (from point-goal navigation) for depth features ($D_t$).
    \item A \textbf{recurrent policy decoder} (GRU) that integrates these inputs to predict the next action.
\end{enumerate}
At each timestep $t$, the GRU updates its hidden state $h_t$ by combining the representations of the current visual inputs and the static instruction. The action $a_t$ is then predicted via a linear layer over the hidden state.
\begin{align}
h_t&=\text{GRU}([V_t, D_t, I], h_{t-1}), \\
a_t&=\mathop{\arg\max}\limits_{a} \text{softmax}(W_a h_t + b_a).
\end{align}

\paragraph{Cross-modal Attention (CMA).}
The CMA model extends the Seq2Seq architecture by incorporating a more sophisticated dual-GRU attention mechanism. Instead of a single recurrent unit, CMA employs two:
\begin{itemize}
    \item A \textbf{first GRU} tracks the observational state. It takes the visual inputs ($V_t, D_t$) and the \textit{previous action} $a_{t-1}$ (as a 32-dimensional embedding) to produce a state $h_t^{1st}$.
    \begin{equation}
        h_t^{1st}=\text{GRU}([V_t, D_t, a_{t-1}],h_{t-1}^{1st}).
    \end{equation}
    \item This state $h_t^{1st}$ is then used as a query in a scaled dot-product attention mechanism to dynamically attend to the instruction features ($\hat{I}_t$), visual features ($\hat{V}_t$), and depth features ($\hat{D}_t$).
    \begin{align}
        \hat{I}_t&=\text{Attn}(I, h_t^{1st}),\\
        \hat{V}_t&=\text{Attn}(V_t,h_t^{1st}),\quad \hat{D}_t=\text{Attn}(D_t,h_t^{1st}).
    \end{align}
    \item A \textbf{second GRU} acts as the decision-making module. It receives a comprehensive set of inputs: the attended features ($\hat{I}_t, \hat{V}_t, \hat{D}_t$), the previous action $a_{t-1}$, and the hidden state of the first GRU $h_t^{1st}$. 
    \item The output of this second GRU, $h_t^{2nd}$, is used to predict the final action $a_t$.
    \begin{align}
        h_t^{2nd} &= \text{GRU}([\hat{I}_t, \hat{V}_t, \hat{D}_t, a_{t-1}, h_t^{1st}], h_{t-1}^{2nd}), \\
        a_t&=\mathop{\arg\max}\limits_{a} \text{softmax}(W_a h_t^{2nd} + b_a).
    \end{align}
\end{itemize}

\paragraph{Hierarchical Neural Radiance (HNR).}
Unlike standard recurrent policies that rely solely on current observations, HNR~\citep{wang2024hnr} introduces a lookahead exploration strategy. It utilizes a pre-trained Hierarchical Neural Radiance representation model to predict (hallucinate) the semantic representations of future environments without physical movement. The framework consists of three key stages:

\begin{itemize}
    \item \textbf{Feature Cloud Encoding:} HNR maintains a feature cloud $\mathcal{M}$ that stores fine-grained visual semantics and 3D spatial information. At step $t$, it extracts grid features $\mathbf{g}_{t,j}$ using CLIP-ViT-B/32 and projects them into 3D space using depth $\mathbf{d}_{t,j}$ and camera intrinsics $\mathbf{K}$:
    \begin{equation}
        P_{t,j} = \left[ \mathbf{d}_{t,j} \mathbf{R}^{-1} \mathbf{K}^{-1} [h, w, 1]^T - \mathbf{T} \right]^T,
    \end{equation}
    where $P_{t,j}$ is the 3D coordinate. The cloud is updated as $\mathcal{M}_{t} = \mathcal{M}_{t-1} \cup \{[\mathbf{g}_{t,j}, P_{t,j}, \theta_{t,j}, s_{t,j}]\}$, including orientation $\theta$ and scale $s$.

    \item \textbf{Future Prediction via Volume Rendering:} To predict the feature map of a future candidate view, HNR employs neural volume rendering. For a sampled point $\mathcal{P}_n$ along a camera ray, the model retrieves the $K$-nearest features from $\mathcal{M}$ using a KD-Tree. To ensure translational invariance, it computes relative positional embeddings:
    \begin{equation}
        \mathbf{q}_k = \text{LN}(\mathbf{W}_{1}[P^{rel}_{k}, \theta^{rel}_{k}, s_{k}]),
    \end{equation}
    where $P^{rel}$ and $\theta^{rel}$ are the relative coordinates and orientation between the query point and the stored features. These are aggregated via an MLP to produce a density $\sigma_n$ and latent vector $\mathbf{r}_n$. The final region feature $\mathbf{R}_{h,w}$ is computed via integration:
    \begin{equation}
        \mathbf{R}_{h,w} = \sum_{n=1}^{N} \tau_n (1 - \exp(-\sigma_n \Delta_n)) \mathbf{r}_n, \quad \tau_n = \exp\left(-\sum_{i=1}^{n-1} \sigma_i \Delta_i\right).
    \end{equation}
    A hierarchical encoder then processes these region features to obtain a holistic view representation.

    \item \textbf{Lookahead Planning:} The predicted future views are organized into a topological graph containing visited nodes, candidate nodes, and lookahead nodes. The graph is encoded using Graph-Aware Self-Attention (GASA), which incorporates the geodesic distance matrix $E$ into the attention mechanism:
    \begin{equation}
        \text{GASA}(\mathcal{V}) = \text{Softmax}\left( \frac{\mathcal{V}\mathbf{W}_q(\mathcal{V}\mathbf{W}_k)^T}{\sqrt{d}} + E\mathbf{W}_e \right) \mathcal{V}\mathbf{W}_v.
    \end{equation}
    Finally, the model evaluates path branches by max-pooling the predicted scores of candidate nodes and their connected lookahead nodes: $S^{path} = \max([S^{candidate}, S^{lookahead}])$.
\end{itemize}

\subsubsection{VLM-based Discrete Planners}
\paragraph{Uni-NaVid.}
Uni-NaVid~\citep{zhang2024uni} reformulates the navigation task as a video-to-text generation problem using a Large Language Model (Vicuna-7B). Given an instruction $\mathcal{I}$ and an ego-centric video history $\mathcal{O}_T=\{\mathbf{x}_1, \dots, \mathbf{x}_T\}$, the model predicts a sequence of $k=4$ future discrete actions. 

To handle long-horizon video efficiently, Uni-NaVid introduces an \textbf{Online Visual Token Merging} mechanism inspired by the Atkinson-Shiffrin memory model. Visual features are first extracted via an EVA-CLIP encoder, $\mathbf{X}_{1:T} = \text{Encoder}(\mathbf{x}_{1:T})$, where each frame yields $N_x$ tokens. These tokens are then categorized into three groups with different compression rates using a Grid Pooling operation:
\begin{equation}
    \mathbf{X}_{1:T} = 
    \begin{cases}
        \mathbf{X}_{\text{curr}} = \text{GridPool}(\mathbf{X}_t, \alpha_\text{curr}), & \text{if } t = T \\
        \mathbf{X}_{\text{short}}= \text{GridPool}(\mathbf{X}_t,  \alpha_\text{short}), & \text{if } t \in [T-B, T) \\
        \mathbf{X}_{\text{long}} = \text{GridPool}(\mathbf{X}_t, \alpha_\text{long}), & \text{if } t \in [1, T-B)
    \end{cases}
\end{equation}
where $\alpha_\text{curr}=2$, $\alpha_\text{short}=8$, and $\alpha_\text{long}=16$ control the spatial pooling resolution, and $B=64$ represents the short-term buffer size.

To enable efficient online inference, the model recursively updates these memory banks. New frames are processed and integrated into the short-term memory, while older short-term tokens are compressed into long-term memory:
\begin{align}
    \mathbf{X}_{\text{curr}\rightarrow\text{short}} &= \text{GridPool}(\mathbf{X}_\text{curr}, \alpha_\text{short}/\alpha_\text{curr}),\\
    \mathbf{X}_{\text{short}\rightarrow\text{long}} &= \text{GridPool}(\mathbf{X}_\text{short}, \alpha_\text{long}/\alpha_\text{short}).
\end{align}
To prevent the long-term memory from growing linearly, Uni-NaVid merges temporally adjacent tokens based on cosine similarity. If the similarity between a candidate token and the existing memory exceeds a threshold $\tau=0.95$, they are fused:
\begin{equation}
    \mathbf{X}_{\text{long}} = \frac{1}{K+1} \left( K \mathbf{X}_{\text{long}} + \mathbf{X}_{\text{short} \to \text{long}} \right) \quad \text{s.t.} \quad \cos(\mathbf{X}_{\text{long}}, \mathbf{X}_{\text{short} \to \text{long}}) > \tau,
\end{equation}
where $K$ is the count of previously merged frames.

Finally, all merged visual tokens are projected via a two-layer MLP, $P_V(\cdot)$, to align with the LLM's embedding space. The input to the LLM is constructed as:
\begin{equation}
    \text{Input} = [\mathbf{E}^V_{\text{long}}, \mathbf{E}^V_{\text{short}}, \mathbf{E}^V_{\text{curr}}, \langle \textit{NAV} \rangle, \mathbf{E}^L_{\text{instr}}],
\end{equation}
where $\langle \textit{NAV} \rangle$ is a task indicator token. The model outputs action tokens autoregressively: $\{\mathbf{E}^A_T, \dots, \mathbf{E}^A_{T+3}\}$.

\subsubsection{Continuous Trajectory Generation Models}
\paragraph{Recurrent Diffusion Policy (RDP).}
Drawing inspiration from the success of diffusion policies in manipulation~\citep{chi2023diffusion} and simpler navigation tasks~\citep{sridhar2024nomad, bar2025navigation, cai2025navdp}, the RDP baseline adapts this generative paradigm to VLN. This approach utilizes a diffusion-based generative head to enable continuous, multi-step trajectory predictions.

The architecture processes ego-centric RGB-D inputs, where RGB images and language instructions are encoded via LongCLIP~\citep{zhang2024long}, while the depth stream is processed by a pre-trained ResNet50. Feature alignment between the visual and linguistic modalities is established through two multi-head, multi-layer cross-modal attention modules~\citep{vaswani2017attention}.

At the core of RDP is a Transformer-based diffusion decoder conditioned on the fused cross-modal features $c_t$. In contrast to discrete planners, RDP predicts a continuous action sequence for the next $T$ steps, denoted as $\{\Delta x_t, \Delta y_t, \Delta{yaw}_t\}_{t=1}^{T}$, corresponding to relative displacement and yaw. The model follows the DDPM framework~\citep{ho2020denoising} for training and sampling, utilizing an iterative denoising process:
\begin{equation}
    a_t^{k-1}=\alpha \cdot (a_t^k-\gamma \epsilon_\theta(c_t, a_t^k, k)+\mathcal{N}(0, \mu^2I)),
\end{equation}
where $k$ represents the denoising step, $\epsilon_\theta$ is the noise prediction network, and $\alpha, \gamma, \mu$ serve as noise schedule functions.

To address the specific challenges of VLN, RDP incorporates two key mechanisms:
\begin{enumerate}
    \item \textbf{History Management:} A recurrent GRU module is employed to maintain and update the observation history, allowing the model to capture long-range dependencies effectively.
    \item \textbf{Stop Prediction:} Since standard diffusion models are not inherently designed for discrete stop/continue decisions, an auxiliary MLP prediction head $\mathcal{S}_{stop}(c_t)$ is added. This head predicts a continuous \textit{stop progression} value $\hat{p}_{stop}$, which evolves from 0 (start) to 1 (goal).
\end{enumerate}
The total loss function combines the standard diffusion denoising objective with the stop progression error:
\begin{equation}
    \mathcal{L}_{\text{RDP}}= \text{MSE}(\epsilon^k, \epsilon_\theta(c_t, a^0_t+\epsilon^k,k)) + \lambda \cdot \text{MSE}(\mathcal{S}_{stop}(c_t), \hat{p}_{stop}),
\end{equation}
where $\hat{p}_{stop}$ represents the ground-truth stop progress and $\lambda$ is a weighting hyperparameter.

\paragraph{Navigation Foundation Model (NFM).}
Proposed by~\citep{zhang2025nfm}, NFM is a generalist navigation policy capable of handling cross-embodiment and cross-task scenarios (\eg, drones, cars, indoor robots). It extends a standard VLM architecture to support multi-view inputs and continuous trajectory prediction.

\begin{itemize}
    \item \textbf{Dual-Scale Visual Encoding:} To capture both semantic semantics and fine-grained details, NFM concatenates features from two pre-trained encoders, DINOv2~\citep{oquab2023dinov2} and SigLIP~\citep{zhai2023sigmoid}. To manage token counts, it applies a Grid Pooling strategy with two resolutions:
    \begin{equation}
        \mathbf{V}^{\text{fine}/\text{coarse}} = \text{GridPool}(\mathbf{V}, s), \quad s \in \{\frac{64}{P}, \frac{4}{P}\},
    \end{equation}
    where fine-grained tokens ($64 \times C$) are used for the most recent observation to ensure precision, while coarse-grained tokens ($4 \times C$) are used for historical frames to save memory.

    \item \textbf{Temporal-Viewpoint Indicator (TVI) Tokens:} To enable the LLM to distinguish between different camera views and time steps in a multi-view setup, NFM adds specialized learnable embeddings to the visual tokens. For a token at time $t$ and view angle $\phi$, the indicator $\mathbf{E}_\text{TVIT}$ is computed as:
    \begin{equation}
        \mathbf{E}_\text{TVIT} = \mathbf{E}_\text{Base} + \mathcal{P}_\text{time}(\text{TimePE}(t)) + \mathcal{P}_\text{angle}(\text{AnglePE}(\phi)),
    \end{equation}
    where $\text{TimePE}$ and $\text{AnglePE}$ are sinusoidal positional encodings processed by MLPs ($\mathcal{P}$). This allows the model to maintain geometric awareness ($0 \equiv 2\pi$) and temporal order simultaneously.

    \item \textbf{Budget-Aware Temporal Sampling (BATS):} To handle long-horizon videos within a fixed token budget $B_\text{token}$, NFM replaces uniform sampling with an exponential decay strategy inspired by the ``forgetting curve." The sampling probability $P(t)$ for a frame at time $t$ relative to the current time $T$ is:
    \begin{equation}
        P(t) = (1-\epsilon)e^{k(t-T)/T} + \epsilon,
    \end{equation}
    where $\epsilon$ is a lower bound and $k$ is a decay rate solved numerically to satisfy the token budget. This ensures high sampling density for recent observations while retaining sparse historical context.

    \item \textbf{Continuous Trajectory Prediction:} Unlike discrete planners, NFM predicts a sequence of continuous waypoints. The LLM outputs a hidden state $E^A_T$, which is decoded by a 3-layer MLP ($\mathcal{A}_\theta$) into normalized coordinates. These are then scaled by a task-specific factor $\alpha_\text{task}$ (\eg, for drone vs. car speeds):
    \begin{equation}
        \tau_T = \{\mathbf{a}_1, \dots, \mathbf{a}_M\} = \alpha_\text{task} \cdot \mathcal{A}_\theta(E^A_T),
    \end{equation}
    where $\mathbf{a} \in \mathbb{R}^4$ represents $(x, y, z, \text{yaw})$.
\end{itemize}

\paragraph{InternNav-N1.}
InternNav-N1~\citep{internnav2025} proposes a dual-system architecture inspired by the ``System 1 vs. System 2" cognitive theory. It decouples navigation into low-frequency reasoning and high-frequency execution to handle long-horizon planning and dynamic obstacle avoidance simultaneously.

\begin{itemize}
    \item \textbf{System 2 (Reasoning Planner):} Built upon a pre-trained VLM (Qwen-VL-2.5 7B~\citep{bai2025qwen2}), System 2 operates at a low frequency (2Hz). It interprets instructions and historical observations to predict a mid-term goal. Initially formulated as a \textit{Pixel Goal} $(u, v)$ on the image plane, this is evolved during joint-tuning into a \textit{Latent Plan}—a set of compact, learnable tokens that bridge the VLM and the policy without geometric ambiguity.

    \item \textbf{System 1 (Agile Executor):} System 1 is a lightweight, multi-modal goal-conditioned diffusion policy running at 30Hz. It takes real-time observations and the asynchronous goal from System 2 to generate continuous trajectories. To ensure robust goal conditioning, System 1 is trained with a \textit{Goal Alignment} objective. Specifically, it encodes both image-goals $I_g$ and pixel-goal masks $M_g$ (derived from coordinates $c_g$) into embeddings $z_{img}$ and $z_{pix}$, enforcing them to align with the ground-truth point-goal $p_g$:
    \begin{equation}
        \mathcal{L}^{goal} = \frac{1}{N}\sum_{i=1}^{N}\| \text{MLP}(z_{img}) - p_g \|^2 + \frac{1}{N}\sum_{i=1}^{N}\| \text{MLP}(z_{pix}) - p_g \|^2.
    \end{equation}

    \item \textbf{Hierarchical Joint Training:} The model employs a two-stage curriculum. First, systems are pre-trained separately using explicit pixel/point goals. Second, they are jointly fine-tuned using latent plans. The overall objective for System 1 combines action generation, critic prediction, and goal alignment:
    \begin{equation}
        \mathcal{L}^{system1} = \alpha \cdot \mathcal{L}^{act} + \beta \cdot \mathcal{L}^{critic} + \gamma \cdot \mathcal{L}^{goal},
    \end{equation}
    where $\mathcal{L}^{act}$ is the diffusion noise prediction loss and $\mathcal{L}^{critic}$ estimates trajectory safety. This asynchronous design allows the agent to perform collision-free traversal while maintaining long-horizon semantic consistency.
\end{itemize}

\subsubsection{Simulator Configuration}
For a fair comparison to previous works, agents in our experiments are set up with the standard dimensions for R2R-CE. Furthermore, while we trained a specialized waypoint predictor using our collected data for our MLLM Zero-shot benchmarks, we employ the original pre-trained waypoint predictor in online-training experiments to strictly align with established baselines. On the other hand, to facilitate the use of the same waypoints predictor and the powerful pre-trained depth-encoder~\citep{wijmans2019dd} in the two benchmarks, as well as to decrease the rendering cost, we adjust the camera parameters in VLNVerse. Some key configurations are listed here, where the commented numbers are the original values.

\begin{tcolorbox}[
    title=\textbf{Online-training Configuration},
    colframe=headergray,      
    colback=white,            
    coltitle=white,           
    fonttitle=\bfseries,
    boxrule=0.5mm,
    arc=1mm,                  
    width=\columnwidth        
]
    \small\ttfamily 
    \begin{minipage}[t]{0.48\textwidth}
        \textbf{SIMULATOR:} \\
        \hspace*{1em} AGENT: \\
        \hspace*{2em} HEIGHT: 1.50 \\
        \hspace*{2em} RADIUS: 0.30 \\
        \hspace*{1em} ISAAC\_SIM: \\
        \hspace*{2em} ENABLE\_COLLISION: True \\
        \hspace*{2em} COLLISION\_THRESH: 0.10 \\
        \hspace*{2em} ALLOW\_SLIDING: True \\
        \hspace*{1em} RGB\_SENSOR: \\
        \hspace*{2em} WIDTH: 224 \\
        \hspace*{2em} HEIGHT: 224 \\
        \hspace*{2em} HFOV: 90 \\
        \hspace*{1em} DEPTH\_SENSOR: \\
        \hspace*{2em} WIDTH: 256 \\
        \hspace*{2em} HEIGHT: 256 \\
        \hspace*{2em} HFOV: 90
    \end{minipage}
    \hfill 
    \begin{minipage}[t]{0.48\textwidth}
        \textbf{TRAINING:} \\
        \hspace*{1em} IL: \\
        \hspace*{2em} BATCH\_SIZE: 4 \\
        \hspace*{2em} MAX\_TRAJ\_LEN: 35 \\
        \hspace*{2em} EPOCHS: 20 \\
        \hspace*{2em} SCHEDULE\_RATIO: 0.85 \\
        \hspace*{2em} DECAY\_TIME: 4 \\
        \hspace*{2em} LR: CosineAnnealingLR \\
        \vspace{0.5em} \\
        \hspace*{1em} ENV: \\
        \hspace*{2em} NUM\_ENVS: 4 \\
        \hspace*{2em} NUM\_GPUS: 4
    \end{minipage}
\end{tcolorbox}

\section{Detailed Metric Definition}
\label{supp:metric}
We evaluate all the models mentioned in Appendix~\ref{supp:models} using a complex set of metrics, the details and definitions can be found in~\cref{tab:metrics_appendix}
\begin{table*}[h!]
\centering
\small 
\caption{Detailed definitions of evaluation metrics.}
\label{tab:metrics_appendix}
\begin{tabularx}{\textwidth}{l X} 
    \toprule
    \textbf{Metric} & \textbf{Detailed Definition} \\
    \midrule

    \textbf{Trajectory Length (TL)} & The average total distance in meters traversed by the agent from its starting position to its final stopping position across all episodes. \\
    \addlinespace

    \textbf{Navigation Error (NE)} & The average distance in meters between the agent's final stopping position and the target goal location. This is calculated for \textit{all} episodes, regardless of their success status. A lower NE indicates better overall navigation accuracy. \\
    \addlinespace
    
    \textbf{Success Rate (SR)} & The percentage of episodes where the agent issues the \texttt{STOP} command within a predefined threshold distance (\eg, 3 meters) of the target goal location. This is a binary metric (1 for success, 0 for failure) calculated for each episode. For long-horizon tasks, this metric is same as SR$_{\text{All}}$. \\
    \addlinespace
    
    \textbf{Oracle Success Rate (OSR)} & The percentage of episodes where the \textit{optimal} path from the agent's \textbf{final stopping position} to the goal is shorter than the success threshold (\eg, 3 meters). This measures "recoverability" and assesses whether the agent stopped in a location from which the goal was realistically reachable. \\
    \addlinespace
    
    \textbf{Success weighted by Path Length (SPL)} & A metric evaluating both task completion and efficiency: $\frac{1}{N} \sum_{i=1}^{N} S_i \frac{L_i}{\max(P_i, L_i)}$, where $N$ is the total episodes, $S_i$ is binary success, $L_i$ is the shortest path length (oracle), and $P_i$ is the agent's path length. This penalizes successful episodes completed via inefficient, long paths. \\
    \addlinespace
    
    \textbf{Normalized Dynamic Time Warping (nDTW)} & A path fidelity metric measuring the similarity between the agent's generated trajectory and the reference (oracle) path. It computes the optimal alignment between the two sequences of 3D points and normalizes by the reference path length. A higher nDTW (closer to 1.0) indicates the agent's trajectory closely matched the shape and structure of the optimal path. \\
    \addlinespace
    
    \textbf{SR$_n$ (Success Rate for $n$-th goal)} & In long-horizon tasks, SR$_n$ measures the \textbf{conditional success rate} of reaching the $n$-th goal. For $n > 1$, this is the percentage of episodes where the agent successfully reached goal $n$, \textit{given} that it had already successfully reached all preceding goals ($1, \dots, n-1$). SR1 measures the non-conditional success rate of the first goal. \\
    \addlinespace
    
    \textbf{Collision Rate (CR)} & A physics-awareness metric defined as the average number of collision events per navigation step. Calculated by dividing the total collisions by the total navigation actions taken. A lower CR indicates a safer agent with better low-level obstacle avoidance. \\
    \bottomrule
\end{tabularx}
\end{table*}
 
\section{A Close Look At VLNVerse}
\label{supp:vis}
In this section, we first visualize the environment to highlight the topological complexity and rendering quality of VLNVerse. Following this, we elaborate on the data generation pipeline, describing the three-agent system (Describer, Verifier, and Synthesizer). We conclude by presenting visualized examples of various task instances constructed using this pipeline.

\subsection{Visualization of VLNVerse Environment}
To complement the quantitative statistics presented in the main text, we provide a qualitative visual analysis of the VLNVerse environments in this section. We showcase the Bird's eye view (BEV) maps and first-person snapshots to empirically demonstrate the structural diversity and high-fidelity rendering that distinguish our dataset from previous works.

\begin{figure}
    \centering
    \includegraphics[width=1.0\linewidth]{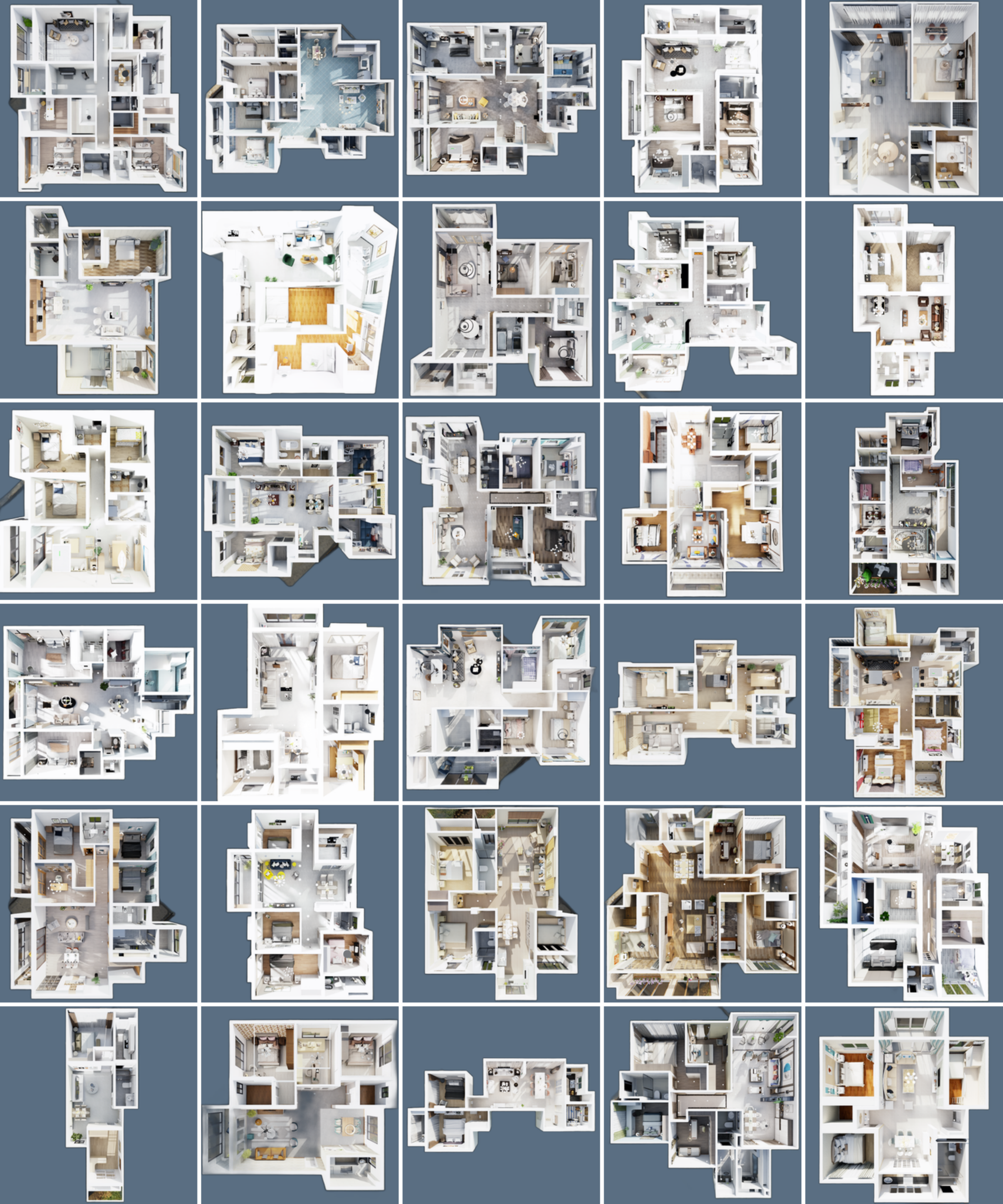}
    \caption{Bird's eye view of VLNVerse environment.}
    \label{fig:scene_bev}
\end{figure}

\paragraph{Structural Diversity.} \cref{fig:scene_bev} displays a curated selection of BEV visualization from our dataset. These maps highlight the topological complexity of our environments. Specifically, VLNVerse features multi-room layouts with varied connectivity graphs. This structural diversity ensures that agents are rigorously tested on long-horizon planning and complex obstacle avoidance.

\begin{figure}
    \centering
    \includegraphics[width=1.0\linewidth]{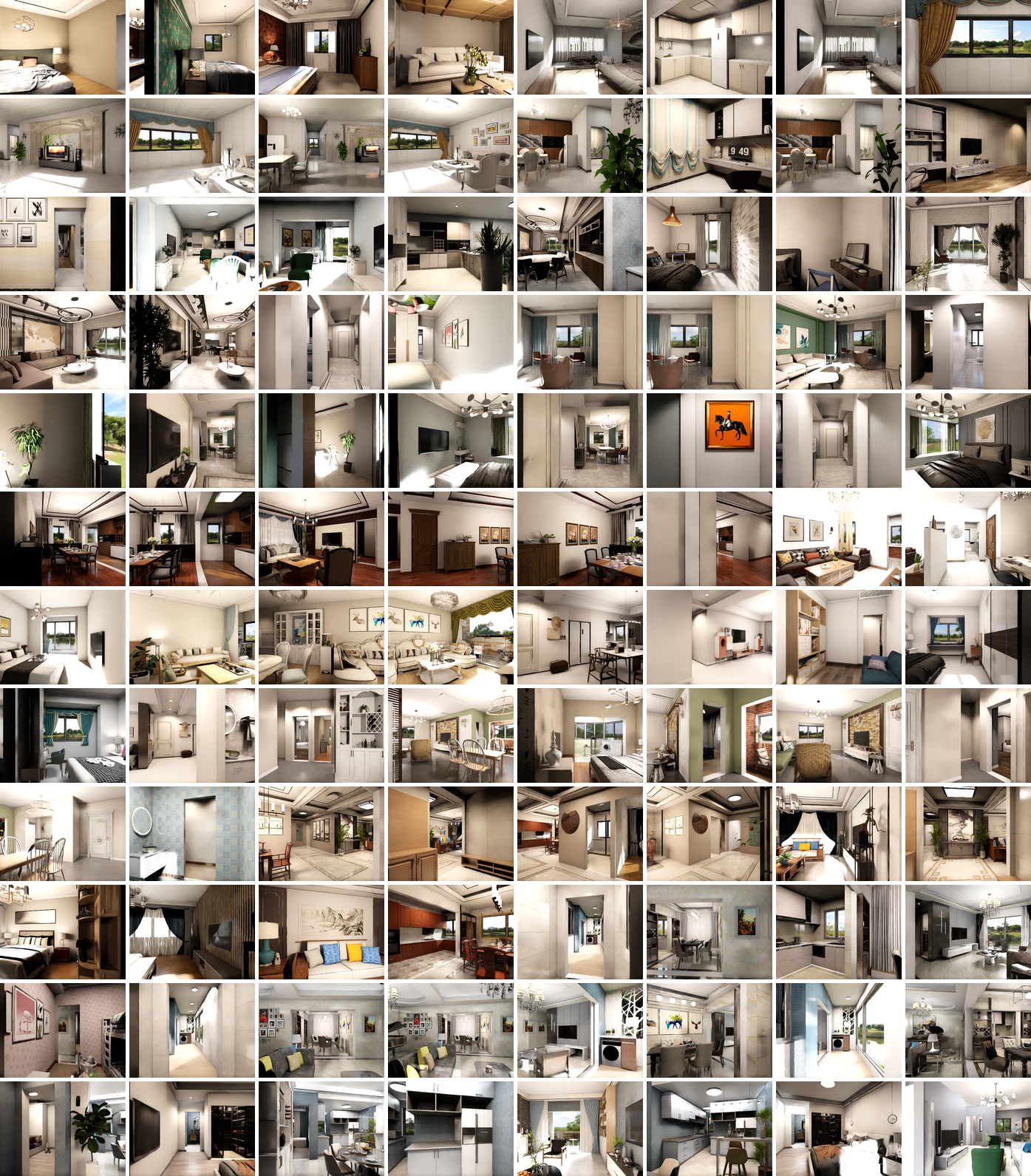}
    \caption{Snapshots of VLNVerse. (Zoom in for details.)}
    \label{fig:snapshot}
\end{figure}

\paragraph{Visual Fidelity.}
\cref{fig:snapshot} presents a gallery of first-person RGB snapshots captured from the agent's perspective. These images demonstrate the photorealistic rendering quality enabled by the NVIDIA Isaac Sim engine and our hand-crafted USD assets. Note the high-resolution textures, realistic lighting effects (\eg, dynamic shadows), and fine-grained object details. Such visual fidelity is crucial for bridging the sim-to-real gap, as it compels the agent's vision encoder to process realistic visual data rather than simplified textures.

\subsection{Data Generation Pipeline}

\subsubsection{Visual Observation}
Based on the collision-free trajectories sampled via A$^*$ on the dilated occupancy maps, we generate high-fidelity visual data using NVIDIA Isaac Sim. To ensure rendering stability and eliminate visual artifacts, we implement a rendering warm-up mechanism, where the simulator performs 20 physics and rendering steps before capturing the final frame at each waypoint. The visual configuration differs slightly between offline data  and online data:
\begin{itemize}
    \item Offline Training Data: We capture the ego-centric front-view RGB images along the trajectory. The camera is configured with a resolution of $224 \times 224$ and a Field of View (FoV) of $90^\circ$. We simultaneously record aligned depth maps of size $256\times 256$, applying a distance filter to clip and normalize pixels exceeding a depth of $10$ meters.
    \item Online Simulation: To support more complex spatial awareness during active navigation, the agent is equipped with a panoptic camera rig. At each step, we capture a full $360^\circ$ view by sampling 12 discrete snapshots at $30^\circ$ yaw intervals, maintaining the same resolution ($224 \times 224$) and FoV ($90^\circ$) as the offline setup.
\end{itemize}

\subsubsection{Navigation Instruction}
Then, we generate the navigation instructions for the sampled paths. We designed a collaborative multi-agent pipeline that fuses structured environment data with natural visual perception. The process follows a Generate-Verify-Synthesize workflow. First, we use the scene-graph prior from the simulator metadata to initialize the factual skeleton. However, while accurate, this prior often lacks visual nuance and relational context. Therefore, we employ a \textbf{Describer} (Agent 1) to analyze the rendered images. To filter out visual hallucinations from the Describer, a \textbf{Verifier} (Agent 2) cross-checks the visual description against the factual prior. Finally, a \textbf{Synthesizer} (Agent 3) aggregates the verified information to generate the final instruction, tailoring the output to specific task requirements and linguistic styles (\eg, formal \vs casual), as follows:

\paragraph{Describer.}
This agent serves as the primary bridge between the visual environment and linguistic instructions. Leveraging a powerful MLLM, the Describer translates raw pixel data into semantic language. Crucially, its behavior is conditioned on the task definition, operating in two distinct modes:

\begin{itemize}
    \item Sequential Description for Fine-grained Navigation:
    For tasks requiring step-by-step guidance, the Describer takes the chronological sequence of egocentric RGB frames from the sampled path as input. We employ a strict prompt engineering strategy to ensure the output is navigable. Specifically, the agent is constrained to use the second-person imperative voice (\eg, ``Turn left", not ``The agent turns left") to mimic natural human commands. Furthermore, the prompt explicitly enforces the inclusion of four key navigational elements: (1) \textit{Landmarks}, (2) \textit{Spatial Information}, (3) \textit{Action Verbs}, and (4) a precise \textit{End Point} description. This ensures the generated text is not merely a video caption, but a functional set of actionable instructions.

    \item Target Profiling for Coarse-grained Navigation:
    For goal-oriented tasks, the Describer shifts focus from the trajectory to the destination state. It analyzes the visual observation of the target object to verify and enrich the scene-graph prior. The agent is prompted to extract specific visual attributes (\eg, color, material, state like ``open/closed") and, critically, to articulate the \textit{spatial relationship} between the target and nearby reference objects (\eg, ``the cup \textit{on} the nightstand"). This ensures that the final instruction (``Find the red cup on the nightstand") is visually grounded and unambiguous.
\end{itemize}

\begin{tcolorbox}[
    colback=blue!5!white,      
    colframe=blue!5!white,     
    arc=2mm,                   
    boxrule=0pt,               
    left=2mm, right=2mm, top=2mm, bottom=2mm 
]
\noindent \textbf{Prompt for Describer (Coarse-grained Instruction)}:
Analyze the provided image, presumably from room [{room\_id}]\footnote{room\_id, target\_object\_name, reference\_object\_name, target\_object\_name are from metadata.}. Focus specifically on the [{target\_object\_name}].
Describe its visual attributes (\eg, color, material if obvious) and its current state (\eg, open/closed, full/empty, on/off if applicable and visible).
Also, describe its spatial relationship to the [{reference\_object\_name}] if visible and relevant. Mention its relationship to any other significant nearby objects as well.
If the target object or reference object isn't clearly visible, the relationship is ambiguous, or the image itself is problematic (\eg, black, blurry), state that clearly (\eg, ``Target object [target\_object\_name] not clearly visible.'' or ``Image is unclear.'').
Respond ONLY with the concise description or the unclear status. Be factual.

\end{tcolorbox}

\begin{tcolorbox}[
    colback=blue!5!white,      
    colframe=blue!5!white,     
    arc=2mm,                   
    boxrule=0pt,               
    left=2mm, right=2mm, top=2mm, bottom=2mm 
]
\noindent \textbf{Prompt for Describer (Fine-grained Instruction)}:
You are an expert navigation assistant. Your goal is to analyze a **sequence of images**, which represents a continuous first-person path, and generate a clear, step-by-step instruction for a person or a robot to follow that exact path. Your highest priority is the accuracy of the actions.

The images are provided in chronological order and should be treated as frames from a video.

The instruction MUST be a direct command written in the second-person imperative voice (\eg, ``Walk forward," ``Turn left"). You are telling an agent what to do.

Your instruction must include these four elements:
1.  **Landmarks:** Refer to areas, rooms, or furniture the agent moves through (\eg, ``enter the living room", ``move towards the bathroom", ``pass by a couch").
2.  **Spatial Information:** Identify the direction of landmarks relative to the agent (\eg, ``with the table on your left", ``the chairs are on the right-hand side").
3.  **Actions:** Detail the agent's movements using command verbs (\eg, ``turn left at the hallway", ``walk straight past the sofa", ``go through the doorway").
4.  **End Point:** Clearly describe the final stopping position (\eg, ``Stop near the sink in front of you", ``Stay in front of the TV", ``Stop to the left of the bed").

**EXAMPLE OF A PERFECT INSTRUCTION:**
``Walk past a dining table on the left and a living room on the right, then turn right into a hallway. Proceed straight down the hallway, and then turn left to enter a bathroom, stopping in front of the window."

**IMPORTANT RULES:**
-   **DO NOT** use third-person descriptive words like ``Walks," ``Moves," ``Enters," or ``Proceeds."
-   **DO NOT** narrate what is happening like a video caption. Give direct commands.
-   Avoid using "Move backward."

Now, generate the navigation instruction for the following image sequence.

\end{tcolorbox}

\paragraph{Verifier.}
While the Describer provides visual richness, it is susceptible to hallucination. Conversely, the scene-graph prior is factually rigid but blind to dynamic visual nuances (\eg, object states). The Verifier acts as a \textit{logic-driven} fusion module to reconcile these two data streams. It receives a batch of rigid, template-based instructions (derived from the prior) alongside the Describer's visual caption. To integrate them, we implement a robust conditional fusion logic:
\begin{itemize}
    \item Visual Priority for Nuance: The agent first evaluates the informativeness of the visual caption. If the vision provides specific, physically plausible spatial relationships or dynamic attributes (\eg, ``the door is \textit{open}", ``the \textit{red} cup") that differ from the generic prior, the agent is explicitly instructed to \textbf{prioritize} the visual evidence. This ensures the instruction reflects the actual rendered reality rather than stale metadata.
    \item Factual Fallback for Safety: If the visual input is deemed ambiguous (\eg, marked as ``unclear" or ``not visible") or describes an implausible relationship, the agent triggers a fallback mechanism, reverting to the spatial relationships defined by the scene-graph Prior. 
\end{itemize}
This two-step logic effectively filters out visual noise while preserving the rich, grounded details necessary for natural instruction generation:
\vspace{-2pt}
\begin{tcolorbox}[
    colback=NavyBlue!5!white,      
    colframe=NavyBlue!5!white,     
    arc=2mm,                   
    boxrule=0pt,               
    left=2mm, right=2mm, top=2mm, bottom=2mm 
]
\vspace{-2pt}
\noindent \textbf{Prompt for Verifier}:
Your task is to analyze 10 rigid text instructions AND an image caption describing the scene. The 10 text instructions share the same semantic goal. The caption provides a targeted description of the goal object, its state/attributes, and its spatial relationships based on an image.**Fusion Rule for Spatial Relationship \& Details:**

- Compare the spatial relationship described in the IMAGE CAPTION (specifically between the target and reference objects) with the one consistently described in the TEXT instructions.

- **IF** the caption is informative (not ``unclear'', ``not visible'', etc.) AND describes a relationship between the correct objects AND this relationship seems physically plausible AND it differs from the text relationship (especially if the text relationship seems implausible): **PRIORITIZE the spatial relationship from the IMAGE CAPTION.** Incorporate relevant attributes/state details from the caption (\eg, color, open/closed) naturally into the generated instructions.

- **ELSE** (caption is uninformative, describes wrong objects, relationship is unclear, or matches the text): **Use the spatial relationship from the TEXT instructions.** You may still incorporate object attributes/state details from the caption if available and relevant. 

**[Text Instructions]** \ {instructions\_text}\footnote{from metadata} \ **[/Text Instructions]**

**[Image Caption (Targeted Description)]** \ {image\_caption}\footnote{from Describer} \ **[/Image Caption]**

\end{tcolorbox}

\paragraph{Synthesizer.}
The final stage of our pipeline is the Synthesizer, responsible for transforming the verified semantic content into diverse natural language. Its primary objective is linguistic style adaptation, ensuring the benchmark covers a wide spectrum of user interaction patterns. The Synthesizer operates differently based on the task granularity:
\begin{itemize}
    \item Coarse-grained Tasks: For coarse-grained navigation, where the user intent can vary significantly in tone, the agent is prompted to generate three distinct variations for every episode:
    (1) \textit{Formal} (precise, objective commands), 
    (2) \textit{Natural} (polite, conversational phrasing), and 
    (3) \textit{Casual} (colloquial, fragmented speech). 
    This stylistic diversity prevents the agent from overfitting to a single linguistic pattern.
    
    \item Fine-grained Tasks: For step-by-step navigation, where clarity and immediacy are crucial, the Synthesizer is restricted to a single natural style. This ensures that the complex instructions remain clear and easy to follow without the ambiguity.
\end{itemize}
Finally, the agent enforces a strict JSON output format to ensure the generated data can be programmatically parsed and validated without manual intervention.

\begin{tcolorbox}[
    colback=TealBlue!5!white,      
    colframe=TealBlue!5!white,     
    arc=2mm,                   
    boxrule=0pt,               
    left=2mm, right=2mm, top=2mm, bottom=2mm 
]
\noindent \textbf{Prompt for Synthesizer}:
You are an expert in natural language generation. 

**Core Task:** Generate three new, high-quality, and distinct instructions (Formal, Natural, Casual) in English based primarily on the TEXT instructions' goal (action, target object, reference object).

**Output Styles:**

1.  **Formal:** Precise, objective, complete sentences. Use details sparingly.

2.  **Natural:** Clear, polite, conversational. Use relevant details naturally.

3.  **Casual:** Very informal, colloquial, uses fragments. Use key details concisely.

You MUST return ONLY a valid JSON object (and nothing else) with the following structure:

{{

  ``formal": ``The formal instruction you generated.",

  ``natural": ``The natural instruction you generated.",

  ``casual": ``The casual instruction you generated."

}}
\end{tcolorbox}

\paragraph{Robustness of Navigation Instruction.} We further assessed the rationality of our three-agent framework by testing it on both GPT-4o and Gemini-2.5-fast. In a blind study, human volunteers were unable to distinguish between instructions generated by different backbones. This underscores the effectiveness of our prompt templates, which successfully guide distinct models to converge on high-quality and fact-based outputs. It demonstrates that our method effectively suppresses model-specific hallucinations by grounding generation in factual priors.

\subsection{Visualization of Task Instances}

\begin{figure}[h!]
    \centering
    \includegraphics[width=1.0\linewidth]{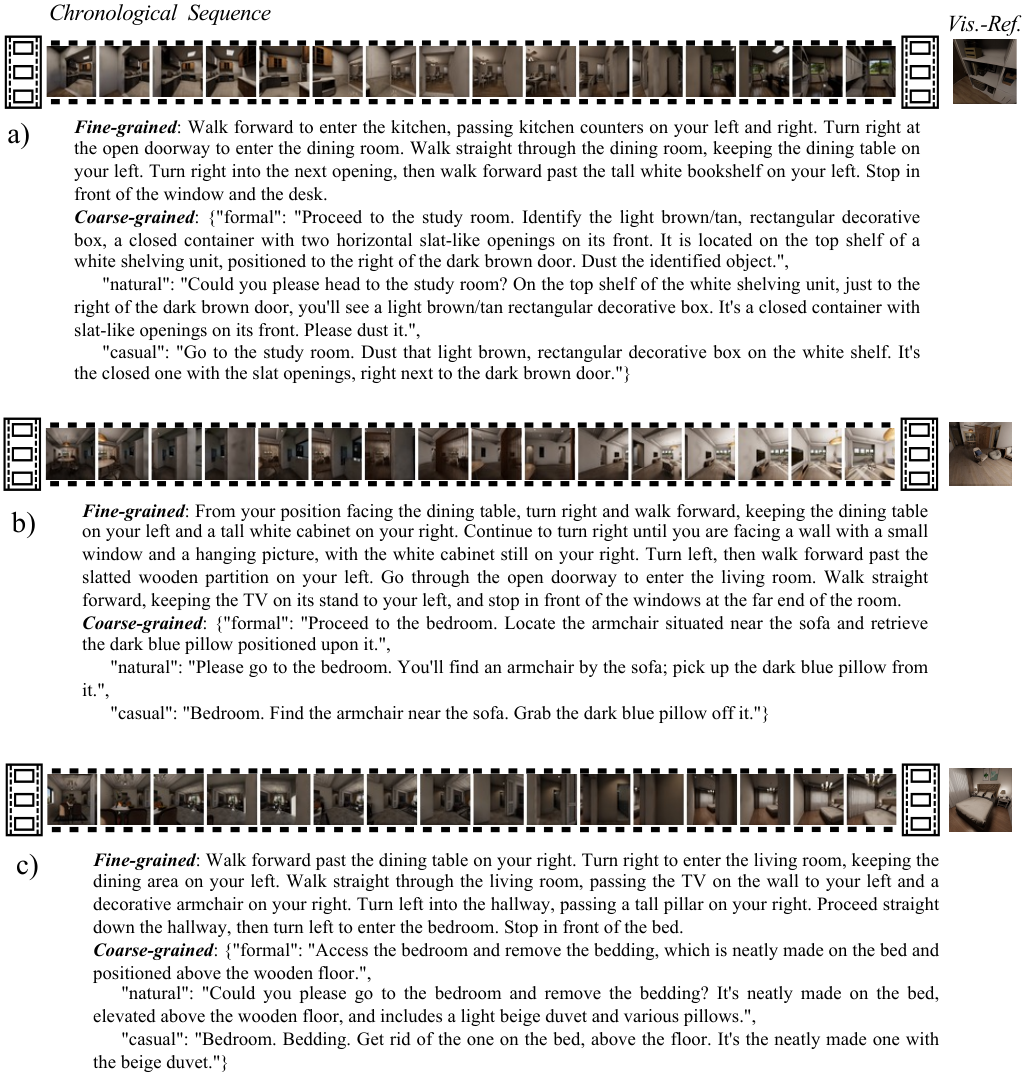}
    \caption{Visualization of sample data representing different task granularities. Each entry consists of a chronological sequence of RGB observations and a visual reference target (\textit{Vis.-Ref.}). (a-c) display three distinct navigation episodes. For each episode, we provide: (1) Fine-grained instructions: Step-by-step navigational guidance describing the path; (2) Coarse-grained instructions: High-level goal descriptions with object interactions, available in three stylistic variations (formal, natural, and casual).}
    \label{fig:data_vis}
\end{figure}

We present qualitative data examples in~\cref{fig:data_vis}. 
Each data entry is constructed around a visual observation sequence (film strips) and paired with instructions at varying granularity. The \textit{Vis.-Ref.} column highlights the modality for visual-reference navigation, providing a cropped target image as the navigational cue. These single-stage episodes serve as the fundamental units for our advanced tasks. We construct long-horizon navigation by chaining these coarse-grained segments, and enable dialogue-based navigation by incorporating an MLLM Oracle (\ie, QwenVL-3~\citep{yang2025qwen3}) for real-time assistance.

\section{Clarification to Human Study}
\label{sec:human_study}

To establish a robust baseline for the proposed tasks, we conducted a comprehensive human study on the test split. This study encompasses both the \textit{Fine-grained} and \textit{Coarse-grained} tasks. Below, we detail the participant selection, experimental interface, data recording mechanism, and ethical considerations.

\paragraph{Participants and Ethics Statement.}
To ensure the results are representative of general users and free from expert bias, we recruited participants with no prior specific knowledge of Vision Language Navigation (VLN). Strict ethical guidelines were followed during the recruitment and testing phases. All participants provided informed consent prior to the study. To ensure fairness and the validity of the evaluation, participants were only exposed to scenes they had not previously encountered, preventing any advantage gained from prior memorization of the environment layout.

\paragraph{Experimental Design and Workload.}
To maintain high attention levels and ensure data quality, we controlled the workload for each tester. Each participant was assigned a single, randomized subset of the testing set, consisting of approximately 50 episodes. On average, participants required between 30 to 60 seconds to evaluate a single episode. This duration was deemed optimal to balance efficient data collection with the prevention of user fatigue.

\paragraph{Interface and Interaction.}
The study was conducted using the Isaac simulator. Upon loading the corresponding data, the navigation instruction was presented to the participant via a pop-up interface. Participants utilized standard keyboard controls to adjust their view and navigate within the 3D environment based on the textual guidance. The termination condition for an episode was user-determined; participants were instructed to close the simulator application once they believed they had successfully reached the target destination described in the instruction.

\paragraph{Data Recording and Evaluation.}
We implemented a high-frequency logging system to capture precise trajectory data. The agent's position and heading were recorded every 50 milliseconds and stored in a CSV file. Performance evaluation was conducted by comparing the recorded human trajectory against the ground-truth path. Since the distribution of episodes was balanced across participants, the final human performance reported in this paper is calculated by averaging the scores of all testers within each respective task (Fine-grained and Coarse-grained).

\begin{figure}
    \centering
    \includegraphics[width=1.0\linewidth]{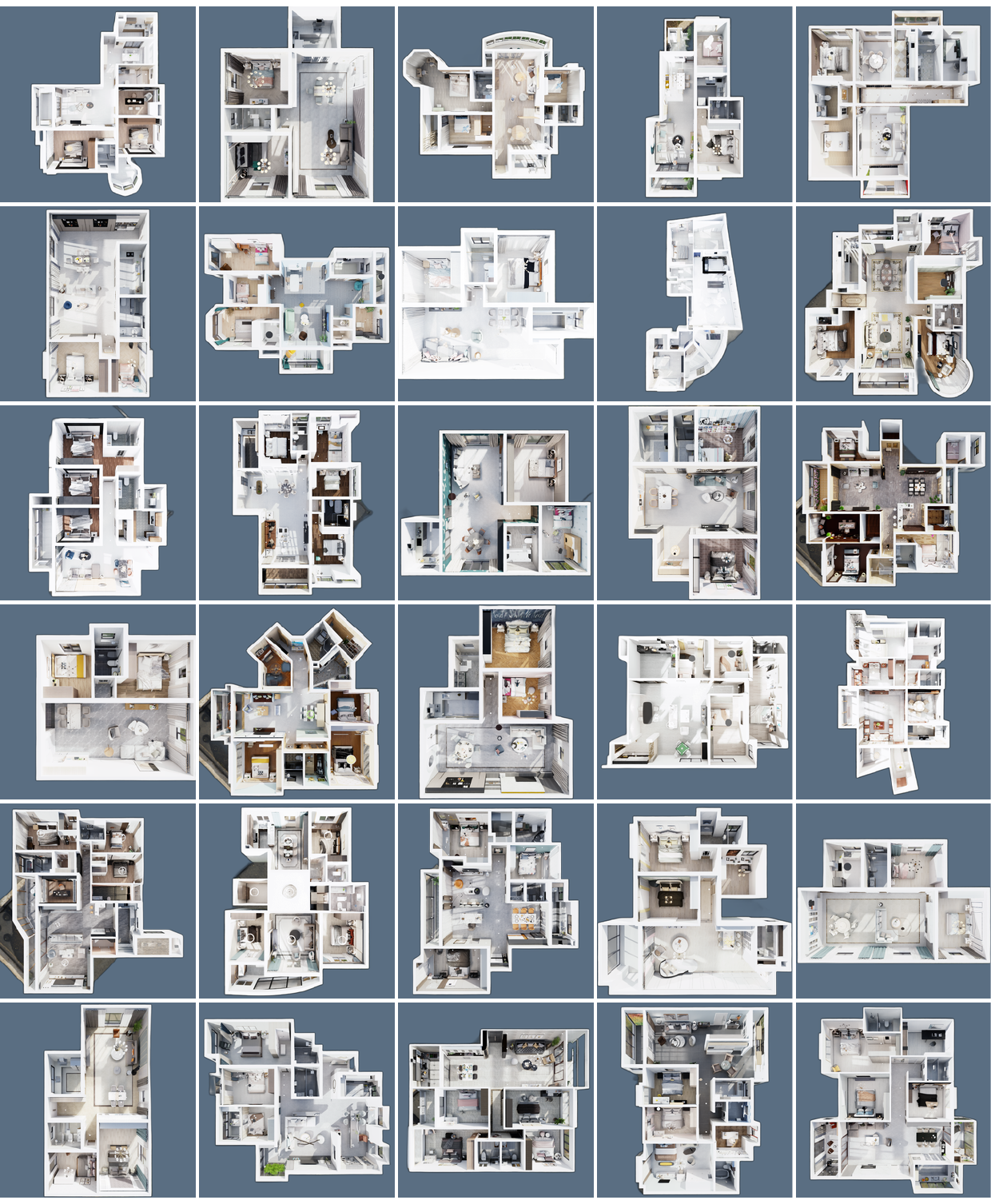}
\end{figure}

\begin{figure}
    \centering
    \includegraphics[width=1.0\linewidth]{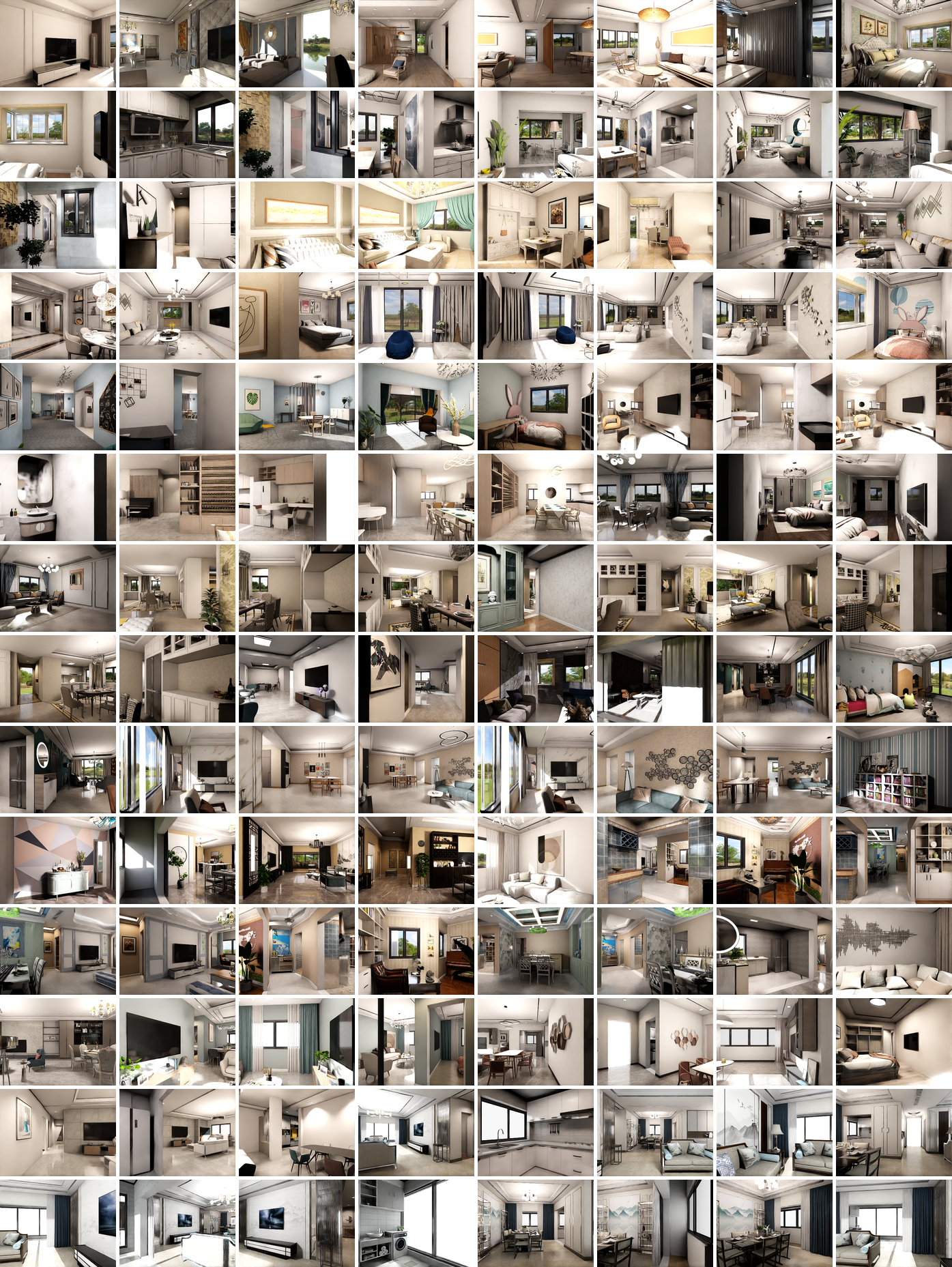}
\end{figure}

\begin{figure}
    \centering
    \includegraphics[width=1.0\linewidth]{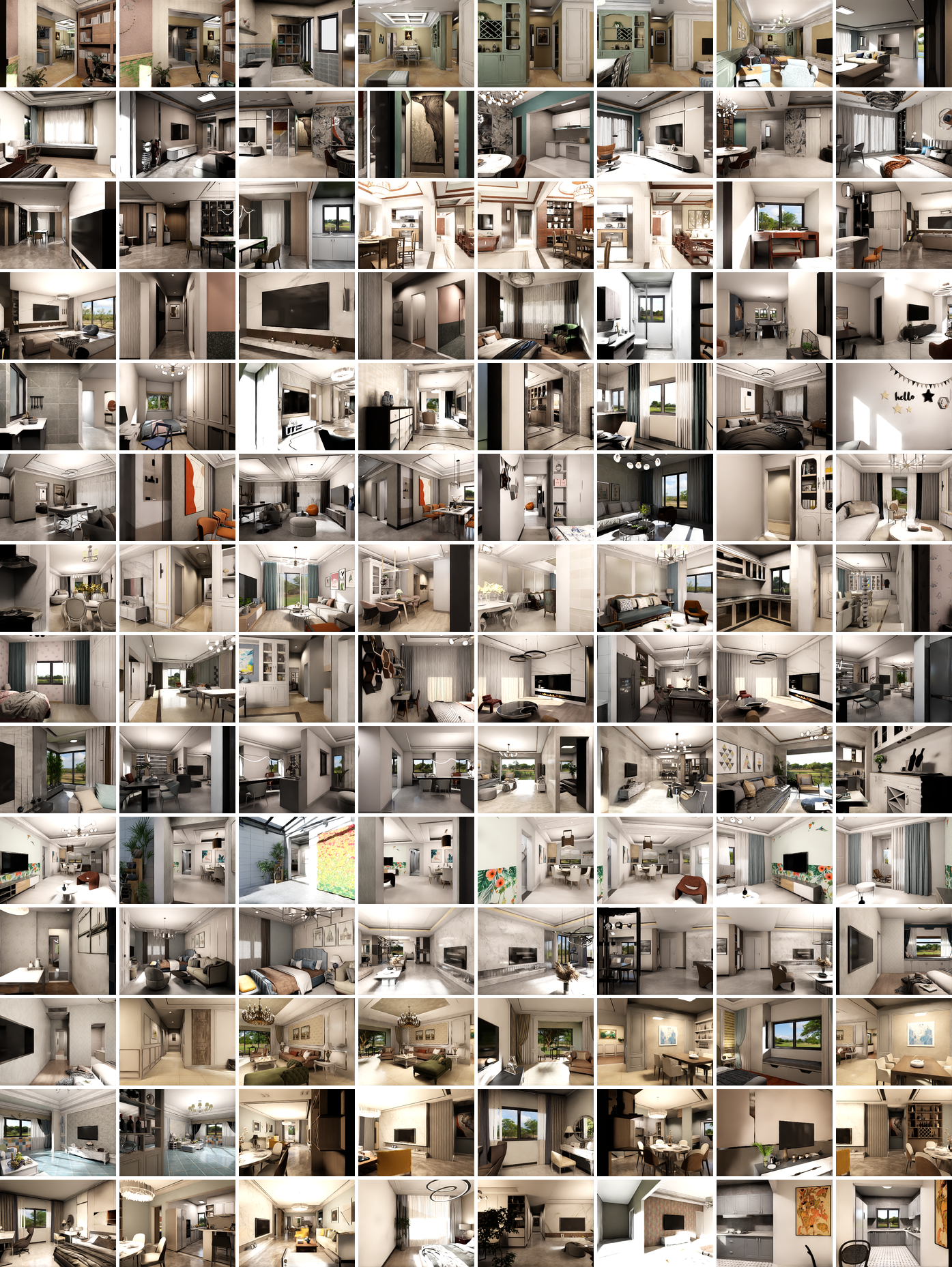}
\end{figure}

\end{document}